\documentclass[10pt,journal,compsoc]{IEEEtran}
\usepackage[pagebackref=false,breaklinks=true, hidelinks,bookmarks=false]{hyperref}
\usepackage{graphicx}
\usepackage{amsmath}
\usepackage{amssymb}
\usepackage{url}
\usepackage{array}
\usepackage{booktabs}
\usepackage{comment}
\usepackage{subcaption}
\usepackage{amssymb}
\usepackage{color}
\usepackage{flushend}
\usepackage{amsmath}
\usepackage{epstopdf}
\usepackage{multirow}
\usepackage{bm}
\usepackage{paralist}
\usepackage{footnote}

\usepackage{hyperref}

\usepackage{ragged2e}

\newcommand{\tabincell}[2]{\begin{tabular}{@{}#1@{}}#2\end{tabular}}

\newcommand{\cosmine}{\operatorname{cos}}
\newcommand{\sinmine}{\operatorname{sin}}

\ifCLASSOPTIONcompsoc
  \usepackage[nocompress]{cite}
\else
  \usepackage{cite}
\fi

\ifCLASSINFOpdf
\else
\fi
\begin{document}
%
\title{View Adaptive Neural Networks for High Performance Skeleton-based Human Action Recognition}

\author{{Pengfei Zhang},
	Cuiling Lan,~\IEEEmembership{Member,~IEEE,}
	Junliang Xing,~\IEEEmembership{Senior Member,~IEEE,} \\
	Wenjun Zeng,~\IEEEmembership{Fellow,~IEEE,}
	Jianru Xue,~\IEEEmembership{Member,~IEEE,}
	Nanning Zheng,~\IEEEmembership{Fellow,~IEEE}

\IEEEcompsocitemizethanks{\IEEEcompsocthanksitem P. Zhang, J. Xue, and N. Zheng are with Xian Jiaotong University, Xian, Shannxi, P. R. China. This work was mostly done while P. Zhang was an intern with Microsoft Research Asia. \protect\\
E-mail: zpengfei@stu.xjtu.edu.cn, \{jrxue,nnzheng\}@mail.xjtu.edu.cn
\IEEEcompsocthanksitem C. Lan, W. Zeng are with Microsoft Research Asia, Beijing, P. R. China. \protect E-mail: \{culan,wezeng\}@microsoft.com
\IEEEcompsocthanksitem J. Xing is with the National Laboratory of Pattern Recognition, Institute of Automatic, Chinese Academy of Sciences, Beijing, P. R. China.  \protect E-mail:jlxing@nlpr.ia.ac.cn}
\thanks{Corresponding authors: Cuiling Lan and Jianru Xue}}




\IEEEtitleabstractindextext{%
\begin{abstract}
\justifying
Skeleton-based human action recognition has recently attracted increasing attention thanks to the accessibility and the popularity of 3D skeleton data. One of the key challenges in action recognition lies in the large variations of action representations when they are captured from different viewpoints. In order to alleviate the effects of view variations, this paper introduces a novel view adaptation scheme, which automatically determines the virtual observation viewpoints over the course of an action in a learning based data driven manner. Instead of re-positioning the skeletons using a fixed human-defined prior criterion, we design two view adaptive neural networks, {\it i.e.,} VA-RNN and VA-CNN, which are respectively built based on the recurrent neural network (RNN) with the Long Short-term Memory (LSTM) and the convolutional neural network (CNN). For each network, a novel view adaptation module learns and determines the most suitable observation viewpoints, and transforms the skeletons to those viewpoints for the end-to-end recognition with a main classification network. Ablation studies find that the proposed view adaptive models are capable of transforming the skeletons of various views to much more consistent virtual viewpoints. Therefore, the models largely eliminate the influence of the viewpoints, enabling the networks to focus on the learning of action-specific features and thus resulting in superior performance. In addition, we design a two-stream scheme (referred to as VA-fusion) that fuses the scores of the two networks to provide the final prediction, obtaining enhanced performance. Moreover, random rotation of skeleton sequences is employed to improve the robustness of view adaptation models and alleviate overfitting during training. Extensive experimental evaluations on five challenging benchmarks demonstrate the effectiveness of the proposed view-adaptive networks and superior performance over state-of-the-art approaches. The source code is available at \textcolor{red}{{\url{https://github.com/microsoft/View-Adaptive-Neural-Networks-for-Skeleton-based-Human-Action-Recognition}}}.
\end{abstract}

\begin{IEEEkeywords}
	View adaptation, skeleton, action recognition, RNN, CNN, consistent
\end{IEEEkeywords}}

\maketitle

\IEEEdisplaynontitleabstractindextext

%
\IEEEpeerreviewmaketitle

\IEEEraisesectionheading{\section{Introduction}\label{sec:introduction}}
\IEEEPARstart{H}{uman} action recognition is an important research area in computer vision with extensive studies and rapid developments in the past decades. It has a wide range of applications, such as visual surveillance, human-computer interaction, video indexing/retrieval, human-computer interaction, game control, video summary and video understanding \cite{IVC10SurveyAction, CVIU11SurveyAction}
\begin{figure}[t] 
	\begin{center}
		\includegraphics[width=1\linewidth]{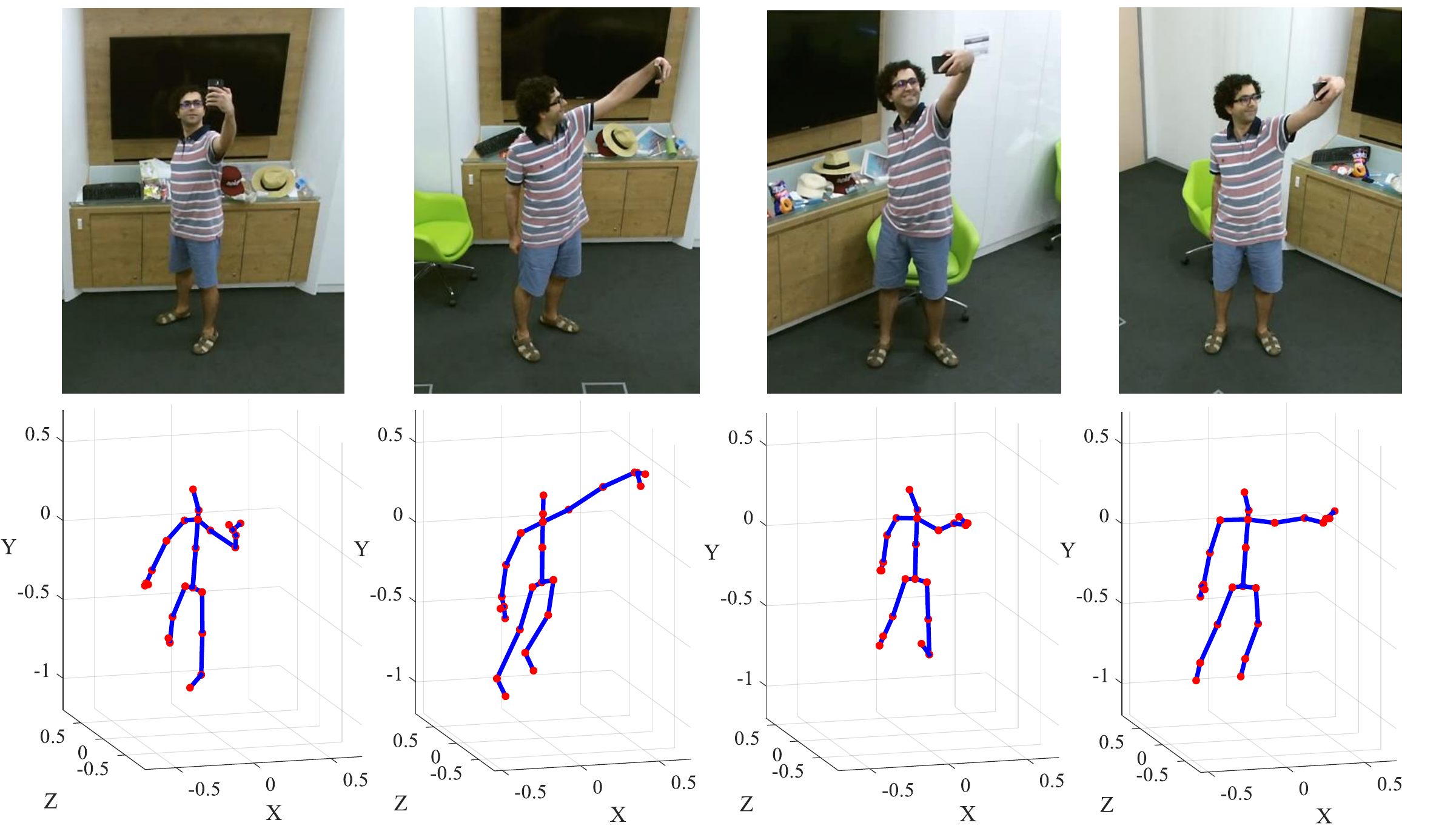}
	\end{center}
	\vspace{-2mm}
	\caption{Skeleton representations of the same posture captured from different viewpoints (different camera position and angle) are very different. } 
	\label{fig:Viewpoints}
\end{figure}
According to the types of input data, human action recognition can be categorized into RGB-based and 3D skeleton-based approaches. RGB-based human action recognition has been studied extensively. 3D skeleton-based human representation, where a human body is represented by the locations of human key joints in the 3D space, has recently attracted increasing attention. It has been demonstrated that RGB-based and skeleton-based approaches for human action recognition complement each other \cite{rahmani2017learning}. As high level representations, skeletons have the merits of being robust to appearances, surrounding distractions, and variations of viewpoints \cite{aggarwal2014human, han2016space, presti20163d, zhang2016rgb}. Biological observations from the early seminal work of Johansson \cite{PP73Perception} suggest that humans are capable of recognizing actions from the motion of just a few joints of the human body, even without appearance information. The prevalence of cost-effective depth cameras such as Microsoft Kinect \cite{zhang2012microsoft}, Intel RealSense \cite{IntelRealSense}, dual camera devices, and the advance of powerful techniques for human pose estimation from depth \cite{CVPR11BestPaper} make 3D skeleton data easy to obtain. Like many previous works discussed in the survey paper \cite{han2016space}, we focus on skeleton-based action recognition in this work.

For human action recognition, one of the main challenges is the large diversity of viewpoints of the captured human action data. There are two major reasons for large view variations. First, in a practical scenario, the viewpoints of the cameras are flexible and different viewpoints result in large differences in skeleton representations even for the same scene. Second, the actor could conduct an action in different orientations. Moreover, he/she may dynamically change his/her orientations as time goes on. As illustrated in Fig. \ref{fig:Viewpoints}, the skeleton representations of the same posture are rather different when captured from different viewpoints. 
In practice, the variation of the observation viewpoints makes action recognition very challenging \cite{aggarwal2014human, ji2010advances}. The viewpoints of the testing samples may have never been seen in those of the training samples and the recognition performance could be significantly degraded. Additionally, to handle diverse viewpoints, a larger model is usually needed in comparison with the case with consistent viewpoints. However, a larger model is prone to be overfitting and is more difficult to train. Some attempts have been made in previous works to overcome the view variation problem for robust action recognition \cite{ji2010advances, rao2001view, bashir2006feature, shen2008ratio, junejo2008selfsimilarities, Farhadi2008Wrongview, shen2009pointtriplets, weinland2010making, liu2011knowtransfer,iosifidis2012ANN, li2012virtualview, wu2012latentSVM, wu2013cross,mahasseni2013latent,zhang2013virtualpath, rahmani2015knotransfer, feng2015usemoreview}. However, most of these works are designed for RGB-based action recognition. They either require training data under many views or their designed view-invariant feature representations do not provide satisfactory performance due to the information loss. For skeleton-based human action recognition, the investigation of view invariance remains under-explored.

There are very few attempts in previous skeleton-based action recognition works which consider the effect of view variations. A pre-processing treatment is usually employed to make the skeleton data invariant to the absolute location and body orientation \cite{CVPR12HO3DJ, vemulapalli2014human, CVPR15HRNN,zhu2015co, jiang2015informative, Shahroudy_2016_CVPR, liu2016spatio,AAAI17Atte}. The original 3D coordinates are transformed to representations under a person-centric coordinate system by placing the body center at the origin and aligning the body plane of skeleton to be parallel to the  $(x, y)$-plane. Such a pre-processing strategy can partially alleviate the view variation problem. However, it introduces some drawbacks. The human-defined strategy may not be flexible enough to handle various cases, considering that human body is not rigid. Besides, those processing methods are not explicitly designed with the target of optimizing action recognition in mind but are instead based on priori knowledge, which reduces the space for exploiting optimal viewpoints. How to design a system which learns optimized viewpoints for action recognition that alleviates the impact of viewpoint diversity without involving much human effort is still an under-explored problem, warranting more investigation.

\begin{figure}[t] 
	\begin{center}
		\includegraphics[width=1\linewidth]{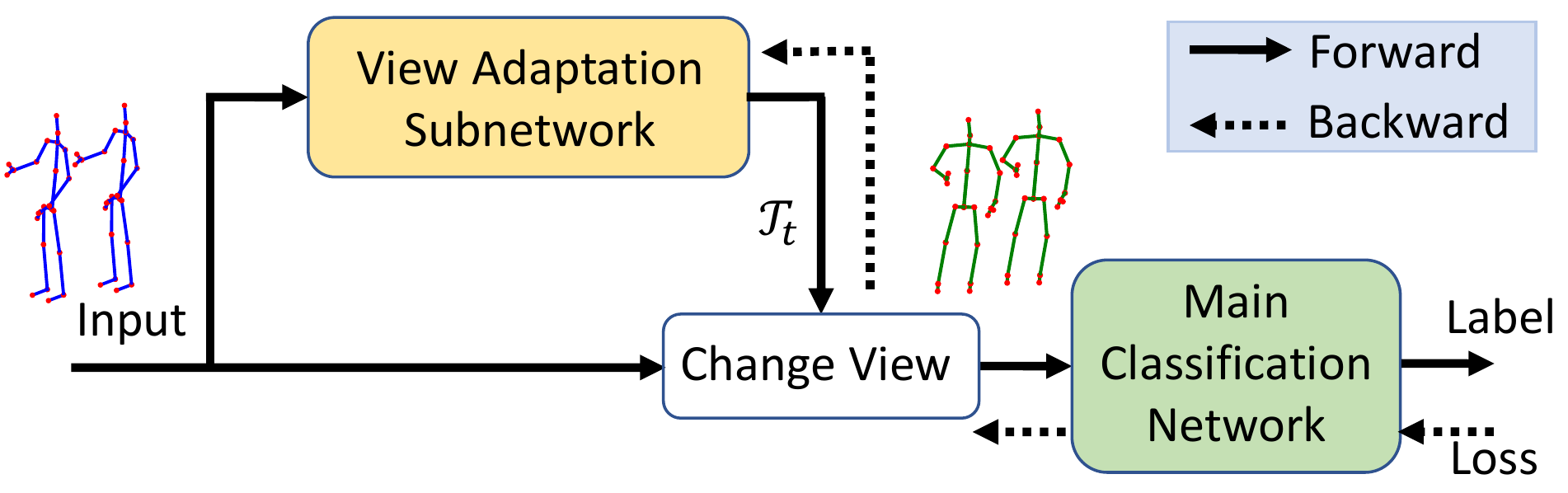}
	\end{center}
	\vspace{-3mm}
	\caption{Flowchart of the end-to-end view adaptive neural network. It consists of a main classification network and a view adaptation subnetwork. The view adaptation subnetwork automatically determines the virtual observation viewpoints and transforms the skeleton input to representations under the new viewpoints for classification by the main classification network. The entire network is end-to-end trained to optimize the classification performance.}
	\label{fig:flowchat}
\end{figure}

In this work, we intend to address the view variation problem to achieve high performance for skeleton-based action recognition. Instead of pre-processing the 3D skeletons based on human defined criteria to reduce view variations, we propose a view adaptation scheme which automatically determines the observation viewpoint within the network for each sample. This enables the classification module to ``see" the skeleton representation under the new viewpoint for efficient recognition. Note that the change of the viewpoint of the camera is equivalent to the transformation of the skeleton to a new coordinate system. As shown in Fig.~\ref{fig:flowchat}, we design an end-to-end view adaptive neural network. It consists of a main classification network and a view adaptation subnetwork.  The view adaptation subnetwork automatically determines the appropriate virtual viewpoints based on the input skeletons. The skeletons represented in the new observation viewpoints are then fed to the main classification network for easier action recognition. With the objective of maximizing the recognition performance, the entire network, including the view adaptation subnetwork and the main classification network, is trained from end to end to encourage the view adaptation subnetwork to learn and determine suitable virtual viewpoints. To demonstrate the effectiveness of our proposed view adaptation mechanism, we apply it to both the recurrent neural network (VA-RNN) and the convolutional neural network (VA-CNN), respectively. One can also fuse the classification scores of these two networks to provide the final prediction , which will be referred to as the two-stream scheme VA-fusion.

Our main contributions are summarized below:
\begin{itemize}
	\setlength{\itemsep}{0pt}
	\item We propose a self-regulated view adaptation scheme which re-positions the observation viewpoints adaptively to facilitate better action recognition from skeleton data. This emancipates the human energy spent on designing complex pre-processing criteria to handle various cases.
	\setlength{\itemsep}{0pt}
	\item We design two view adaptive neural networks, VA-RNN and VA-CNN. For the VA-RNN, we integrate an RNN-based view adaptation module into an LSTM classification network for end-to-end learning. For the VA-CNN, we integrate a CNN-based view adaptation module into a CNN classification network for end-to-end learning. In each stream, the view adaptation module automatically determines the ``best" observation viewpoints during recognition. 
	\setlength{\itemsep}{0pt}	
	\item We perform extensive ablation studies. The effectiveness of the view adaptation subnetworks is demonstrated in extensive experiments. The influence of parameters is analyzed. Moreover, we demonstrate the gain is brought by our view adaptation module rather than the simple increase of layers. We visualize the transformed skeletons from both VA-RNN and VA-CNN networks to better understand why our models work well. Failure cases are discussed. 
	\item To enhance the ``power" of our view adaptation subnetworks, view enriching by data augmentation is performed by randomly rotating the skeletons during training. Experimental results demonstrate that data augmentation improves the robustness of our view adaptation models.

	\setlength{\itemsep}{0pt}

\end{itemize}

Based on the above model innovations and technical contributions, we have presented a very effective skeleton based action recognition system which automatically regulates the skeletons to more consistent viewpoints while maintaining the continuity of an action. 

It should be mentioned that this paper is an extension of our previous conference paper \cite{zhang2017VA}. As an extension, we validate the effectiveness of the view adaptation schemes on both the recurrent neural network, and convolutional neural network. A two-stream scheme by fusing the scores of the two view adaptive networks provides much better performance. We introduce view enriching on the samples during training to further enhance the robustness of the view adaptation model to the view variations. We also conduct the extensive experimental analysis on the network designs in terms of the parameter determinations. Moreover, we conduct experiments on five challenging datasets and experimental results show that our proposed scheme consistently achieves significant gains on the various datasets, demonstrating the effectiveness of our proposed models. 

\begin{figure*}[htbp] 
	\begin{center}
		\includegraphics[width=0.85\linewidth]{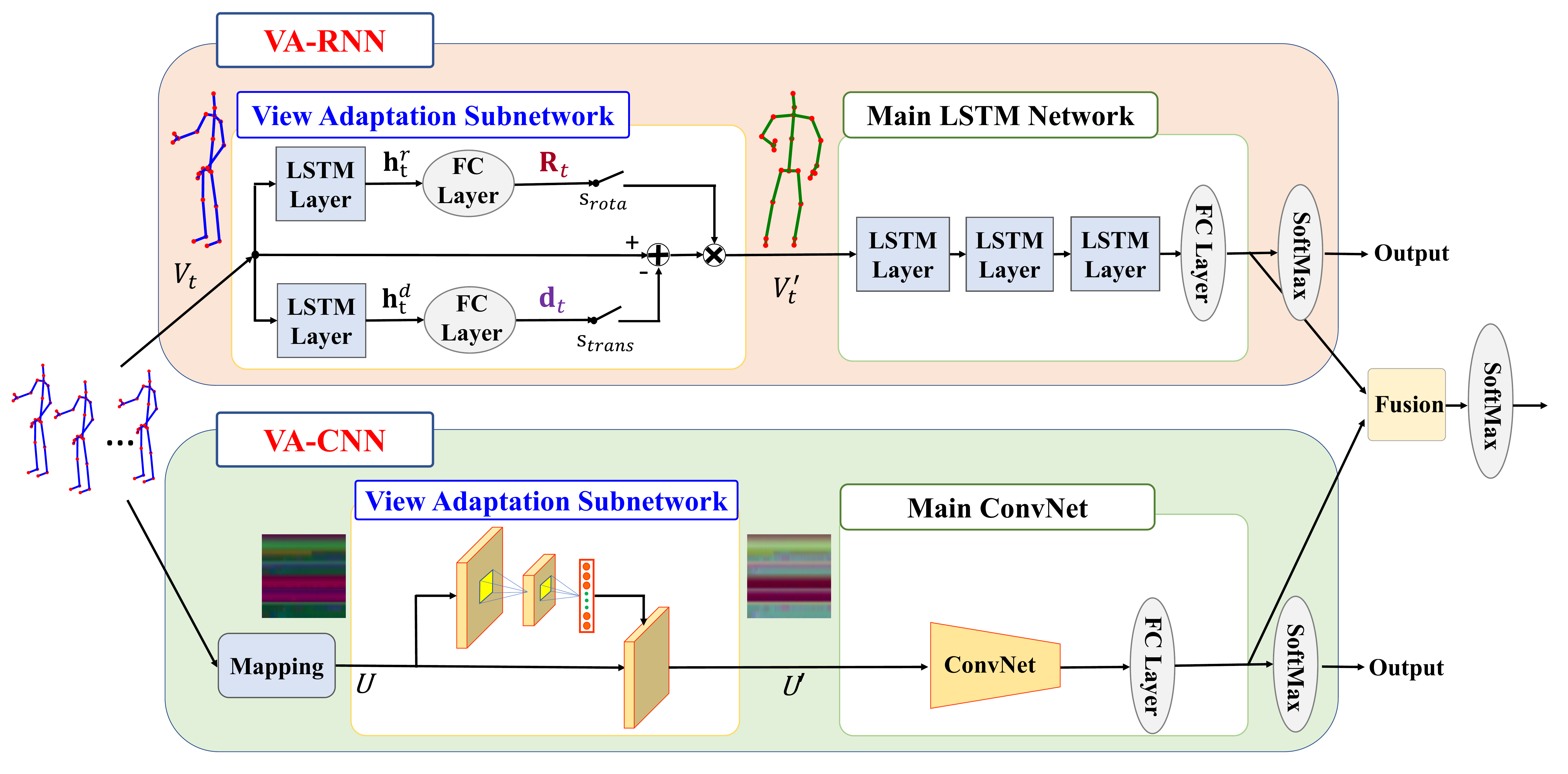} 
	\end{center}
	\vspace{-2mm}
	\caption{Architecture of the proposed view adaptive neural networks: a view adaptive RNN with LSTM (VA-RNN), and a view adaptive CNN (VA-CNN). The VA-RNN consists of a view adaptation subnetwork, and a main LSTM network. The view adaptation subnetwork determines the suitable observation viewpoint at each time slot. With the skeleton representations under new observation viewpoints, the main LSTM network determines the action class. The VA-CNN consists of a view adaptation subnetwork, and a main convolutional network (ConvNet). The view adaptation subnetwork determines the suitable observation viewpoints for the sequence. With the skeleton representations under the new observation viewpoint, the main ConvNet determines the action class. The classification scores from the two networks can be fused to provide the fused prediction, denoted as the VA-fusion scheme.}
	\label{fig:framework}
\end{figure*} 
\section{Related Work}
\label{sec:relatedwork}

\subsection{View Invariant RGB-based Action Recognition}
\label{subsec:VI-RGB}
In reality, human actions can be captured from arbitrary camera viewpoints. This makes it difficult to develop efficient action recognition techniques. For RGB-based action recognition, researchers have paid much attention to this issue and proposed various view-invariant approaches \cite{ji2010advances,rao2001view,bashir2006feature,shen2008ratio,junejo2008selfsimilarities, Farhadi2008Wrongview, shen2009pointtriplets, weinland2010making, liu2011knowtransfer,iosifidis2012ANN, li2012virtualview,wu2012latentSVM, wu2013cross,mahasseni2013latent,zhang2013virtualpath, rahmani2015knotransfer,feng2015usemoreview}.
One category of approaches requires as many views as possible for training to get a `panorama' model \cite{iosifidis2012ANN,feng2015usemoreview,weinland2010making,wu2013cross,mahasseni2013latent}. For example, the 3D histogram of Oriented Gradients based Bag of Words model \cite{weinland2010making} is learned from data of all viewpoints to provide robustness to view changes. In practice, it is expensive to capture videos from abundant views. Another category of approaches designs view-invariant feature representations  \cite{junejo2008selfsimilarities,rao2001view,bashir2006feature} like self-similarity descriptors \cite{junejo2008selfsimilarities} or descriptions based on trajectory curvature \cite{rao2001view,bashir2006feature}. Usually, the descriptors, which are presented in another domain, lose some information of the original video sequences. There is also a category of  approaches that employs knowledge transfer-based models \cite{Farhadi2008Wrongview,liu2011knowtransfer,li2012virtualview,zhang2013virtualpath,zheng2013sparse,rahmani2015knotransfer}. They find a view independent latent space in which features from different views are directly comparable. It requires significant human effort to find the view independent latent space in the design. Considering the different domains of color videos and skeleton sequences, the approaches designed for color videos cannot be directly extended to skeleton-based action recognition.  

\subsection{Viewpoints in Skeleton-based Action Recognition}
For skeleton-based action recognition, the study of viewpoint influences is under-explored. To be view-invariant, the commonly used strategies are to perform pre-processing of skeletons \cite{CVPR12HO3DJ,vemulapalli2014human,CVPR15HRNN,zhu2015co,jiang2015informative, Shahroudy_2016_CVPR, liu2016spatio,AAAI17Atte,liu2017global,li2017RNNTree}. Unfortunately, frame-level pre-processing, where each frame is transformed to the body center with the upper-body orientation aligned, usually results in the partial loss of  relative motion information. For example, the action of walking becomes walking in the same place and the action of dancing with body rotating becomes dancing with body facing a fixed orientation. Sequence-level pre-processing performs the same transformation on all frames with the same parameters determined from the first frame, in which case the motion is invariant to the initial body position and orientation ,and the motion information is preserved. However, since the human body is non-rigid, the definition of the body plane by the joints of ``hip", ``shoulder", ``neck" is not always suitable for the purpose of orientation alignment \cite {wang2014learning}. For example, after the alignment of such a defined body plane, a person who is bending over will have his/her legs upward. Wang {\it et al.} \cite{wang2014learning} use only the up-right pose frames in a sequence to determine the body plane by averaging the rotation transformation. However, a sequence may not contain an up-right pose.  

Rather than struggling to define complex criteria to handle the non-rigid
body, we leverage content-dependent view adaptation models to automatically learn and determine the suitable viewpoints, and transform the skeletons to  representations under those views for each sequence.

\subsection{RNN for Skeleton-based Action Recognition}

In earlier works, hand-crafted features are used for action recognition from the skeleton \cite{han2016space,CVPR12HO3DJ}. With the advance of deep learning, many recent works leverage Recurrent Neural Networks to recognize human actions from raw skeleton input, where the feature learning and temporal dynamic modeling are accomplished by the neural networks. Du {\it et al.} \cite{CVPR15HRNN} propose an end-to-end hierarchical RNN for action recognition. It splits the whole human body into five parts with each part fed into different subnetworks and the output of the subnetworks are fused hierarchically. In the part aware LSTM model \cite{Shahroudy_2016_CVPR}, Amir {\it et al.} propose to split a cell in an LSTM into part-based cells, with each cell learning long-term context representations for each individual part rather than for the entire body. Zhu {\it et al.} \cite{zhu2015co} propose to automatically explore the co-occurrence of discriminative skeleton joints in an LSTM network based on group sparse regularization. To explicitly explore both the spatial and temporal relationships among joints, the spatial-temporal LSTM network extends the LSTM architecture to two concurrent domains, {\it i.e.}, the temporal domain and the spatial domain \cite{liu2016spatio}. A trust gate is introduced to reduce the influence of noisy joints. To further exploit the discriminative powers of different joints and frames, a spatial-temporal attention model \cite{AAAI17Atte} further introduces the attention mechanism into the network to enable it to selectively focus on discriminative joints of the skeleton within one frame, and pay different levels of attention to the outputs at different time instances. Similarly, Liu {\it et al.} \cite{liu2017global} use both global contextual information and local information to selectively focus on  informative joints. Li {\it et al.} \cite{li2017RNNTree} combine tree-like hierarchy RNNs with action category hierarchy to distinguish easy-tell actions in the low levels of networks and hard-tell actions in the high levels of networks.

Different from the above works, we enhance the recognition performance from a new perspective. We leverage RNNs to automatically determine the virtual observation viewpoints and thus the new skeleton representations for efficient action recognition. Note that most previous works take the center and orientation aligned skeletons as input to the RNNs, by using the human defined alignment criteria which are not flexible and not effective.

\subsection{ConvNet for Skeleton-based Action Recognition}	
Considering the strong capability of convolutional neural networks for classification, several recent works \cite{du2015skeleton,wang2016action, liu2017enhanced, li2017skeleton, ke2017new} attempt to convert a skeleton sequence to 2D image(s) and then leverage convolutional neural networks for classification. Some works \cite{du2015skeleton, li2017skeleton} assign the values of the $x$, $y$ and $z$ axes in 3D coordinates to three channels of an image, with the frame ids corresponding to different rows (or columns) and joint ids corresponding to different columns (or rows). Normalization of the coordinate values to the range of 0-255 is performed based on dataset statistics \cite{du2015skeleton} or sequence statistics \cite{li2017skeleton}. Instead of using the absolute coordinate values, He {\it et al.} \cite{ke2017new} use the relative positions between the joints and the reference joints ({\it e.g.} left shoulder, right shoulder, left hip and right hip) to generate multiple images. Some works \cite{wang2016action,liu2017enhanced} use the 2D projection maps from the trajectories of joints to different orthogonal planes as the images. Liu {\it et al.} \cite{liu2017enhanced} represent a 5D space (3D coordinates, time label, joint label) as a 2D coordinate space and a 3D color space. Thus 10 images obtained from different assignments of the 5D space are fed to 10 ConvNets for classification, respectively. The results from the 10 models are fused as the final prediction. 
		
Most of the above works focus on how to map a skeleton sequence to image(s) but ignore the challenge caused by view variations of the skeleton data. In contrast, we leverage ConvNets to automatically determine the virtual observation viewpoints and thus the new skeleton maps for efficient CNN-based action recognition. 

\section{View Adaptation Modeling}
\label{sec:proposed}

The skeleton representations under different views are very different even for the same action. The intra-class differences caused by the view variations may be  even larger than inter-class differences. The diversity of capturing viewpoints makes human action recognition challenging. 

To eliminate the influences of viewpoints, as illustrated in Fig. \ref{fig:flowchat}, we propose an end-to-end neural network architecture which automatically re-observes a skeleton sequence from new virtual viewpoints before action recognition. It consists of a main classification network and a view adaptation subnetwork. The view adaptation module automatically determines the virtual observation viewpoints and outputs a set of transform parameters $\mathcal{T}_t$ for each time $t$ (or $\mathcal{T}$ for a sequence). The input skeleton representation is transformed to representations under the new viewpoints for classification by the main classification network. The entire network is end-to-end trained to optimize the classification performance. 

In the following subsection, we formulate the problem of observation viewpoint adaptation transformation, which is the role played by the view adaptation module.  

\subsection{Problem Formulation}
\label{subsec:formulation}
Captured raw 3D skeletons are representations corresponding to the camera coordinate system (global coordinate system), with the coordinate origin located at the position of the camera sensor. To be insensitive to the initial position of an action and to facilitate our study, for a sequence, we translate the global coordinate system to the body center of the first frame as our new global coordinate system $\mathcal{O}$. Note that the input skeleton $V_{t}$ for our system as illustrated in Fig.~\ref{fig:framework} is the skeleton representation under the new global coordinate system. 
 
\begin{figure}[t] 
	\begin{center}
		\includegraphics[width=1\linewidth]{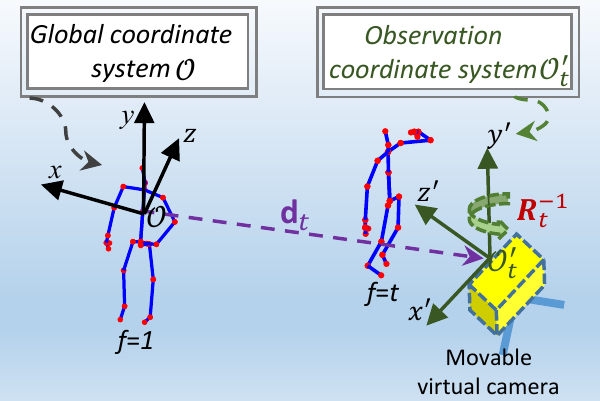}
	\end{center}
	\vspace{-2mm}
	\caption{Illustration of the change of the observation viewpoint, by assuming there is a movable virtual camera. A skeleton sequence is a record of the skeletons from the first frame $f\!=\!1$ to the last frame $f\!=\!T$ under the global coordinate system $\mathcal{O}$. The action can be re-observed by a movable virtual camera under the observation coordinate systems. For the $t^{th}$ frame, the observation coordinate system is at a new position which is obtained from a translation of $\mathbf{d}_t$ and a rotation of $\alpha_t$, $\beta_t$, $\gamma_t$ radians anticlockwise around the $X$-axis, $Y$-axis, and $Z$-axis, respectively, corresponding to the global coordinate system. The skeleton can then be represented under this observation coordinate system $\mathcal{O}'_t$.}
	\label{fig:viewRegulation}
\end{figure}

In TV or movie shooting, one can choose to observe an action from suitable viewpoints over time to better understand the scene and better tell a story. Thanks to the availability of the 3D skeletons captured from a certain view, it is similarly possible to set up a movable virtual camera and observe the action from new observation viewpoints as illustrated in Fig.~\ref{fig:viewRegulation}. With the skeleton at frame $t$ re-observed from the movable virtual camera viewpoint (observation viewpoint), the skeleton is transformed to a representation under the movable virtual camera coordinate system, which is also referred to as the observation coordinate system $\mathcal{O}'_t$. 

Given a skeleton sequence $\mathcal{S}$ under the global coordinate system $\mathcal{O}$, the $j^{th}$ skeleton joint on the $t^{th}$ frame is denoted as $\mathbf{v}_{t,j} = [x_{t,j},y_{t,j},z_{t,j}]^{\rm T}$, where $t \in (1,\cdots, T)$, $j \in (1,\cdots, J)$, $T$ denotes the total number of frames in a sequence, $J$ denotes the total number of skeleton joints in a frame. The set of joints in the $t^{th}$ frame is denoted as $V_t = \{\mathbf{v}_{t,1}, \cdots, \mathbf{v}_{t,J}\}$. 

For the $t^{th}$ frame, we assume the movable virtual camera is placed at a suitable viewpoint, with the corresponding observation coordinate system obtained from a translation by $\mathbf{d}_t \in \mathbb{R}^3$, and a rotation of $\alpha_t$, $\beta_t$, $\gamma_t$ radians anticlockwise around the $X$-axis, $Y$-axis, and $Z$-axis,  respectively, of the global coordinate system. We denote the set of transformation parameters as $\mathcal{T}_t=\{\alpha_t, \beta_t, \gamma_t, \mathbf{d}_t\}$. Therefore, the representation of the $j^{th}$ skeleton joint $\mathbf{v}'_{t,j} = [x'_{t,j},y'_{t,j},z'_{t,j}]^{\rm T}$ of the $t^{th}$ frame under the observation coordinate system $\mathcal{O}'_t$ is 
\begin{equation}
\label{equ:transform}
\mathbf{v}'_{t,j} = [x'_{t,j},y'_{t,j},z'_{t,j}]^{\rm T} = \mathbf{R}_t (\mathbf{v}_{t,j} - \mathbf{d}_t).
\end{equation}
$\mathbf{R}_t $ is represented as 
\begin{equation}
\label{equ:Rxyz}
\mathbf{R}_t = \mathbf{R}_{t,\alpha}^x  \mathbf{R}_{t,\beta}^y \mathbf{R}_{t,\gamma}^z,
\end{equation}
where $\mathbf{R}_{t,\alpha}^x$, $\mathbf{R}_{t,\beta}^y$,  $\mathbf{R}_{t,\gamma}^z$, denote the coordinate transformation matrixes for rotating the original coordinate system around the $X$-axis by $\alpha_t$ radians, the $Y$-axis by $\beta_t$ radians, and the $Z$-axis by $\gamma_t$ radians anticlockwise, respectively,  which are defined as

\begin{equation}
\label{equ:rotation_x}
\mathbf{R}_{t, \alpha}^x = \begin{bmatrix}
1 & 0 & 0\\
0 &\cosmine(\alpha_t) & \sinmine(\alpha_t)  \\
0 & -\sinmine(\alpha_t) & \cosmine(\alpha_t) \\
\end{bmatrix},
\end{equation}

\begin{equation}
\label{equ:rotation_z}
\mathbf{R}_{t,\beta}^y = \begin{bmatrix}
\cosmine(\beta_t) & \sinmine(\beta_t) & 0 \\
-\sinmine(\beta_t) & \cosmine(\beta_t) & 0 \\
0 & 0 & 1
\end{bmatrix},
\end{equation}

\begin{equation}
\label{equ:rotation_y}
\mathbf{R}_{t,\gamma}^z = \begin{bmatrix}
\cosmine(\gamma_t) & 0 & -\sinmine(\gamma_t) \\
0 & 1 & 0 \\
\sinmine(\gamma_t) & 0 & \cosmine(\gamma_t)
\end{bmatrix}.
\end{equation}
Note that all the skeleton joints in the $t^{th}$ frame share the same transformation parameters, {\it i.e.}, $\alpha_t, \beta_t, \gamma_t, \mathbf{d}_t$. This is because changing viewpoints is a rigid motion. Given these transformation parameters, the skeleton representation $V'_t = \{\mathbf{v}'_{t,1}, \cdots, \mathbf{v}'_{t,J}\}$ under the new observation coordinate can be obtained from (\ref{equ:transform}). The viewpoints can vary over time for different frames. The key problem becomes how to determine the viewpoints of the movable virtual camera.

\section{View Adaptive Neural Networks} 
\label{sec:two-stream}
We design two view adaptive neural networks based on RNN and CNN, named VA-RNN  and VA-CNN, respectively. As illustrated in Fig.~\ref{fig:framework}, the VA-RNN (as shown on the top),  consists of a RNN-based view adaptation subnetwork for transforming the skeletons to new representations under the suitable observation viewpoints, and a main LSTM network for recognizing actions from the transformed skeletons. The VA-CNN (as shown on the bottom) consists of a CNN-based view adaptation subnetwork, and a main convolutional network (ConvNet). Each network is trained end-to-end by optimizing the classification performance. Alternatively, we can fuse the scores from the two networks to provide a fused prediction, denoted as the VA-funsion scheme. 

\subsection{View Adaptive Recurrent Neural Network (VA-RNN)}
\label{subsec:network}

As illustrated by the top branch in Fig.~\ref{fig:framework} as marked by VA-RNN, we use a view adaptation subnetwork to automatically learn and determine the observation viewpoints, {\it i.e.}, with transformation parameters of $\alpha_t, \beta_t, \gamma_t, \mathbf{d}$ (as discussed in section \ref{subsec:formulation}), and use a main LSTM network to learn the temporal dynamics and features from the view-adapted skeleton data for action recognition, from end to end.

\textbf{View Adaptation Subnetwork.} The adaptation of observation viewpoint can be considered as the re-positioning of the movable virtual camera, which is characterized by the translation and rotation of this virtual camera (observation coordination system). At a time slot corresponding to the $t^{th}$ frame, with a skeleton $V_{t}$ as input, two branches of LSTM subnetworks are utilized to learn the rotation parameters $\alpha_t, \beta_t, \gamma_t$ to obtain the rotation matrix $\mathbf{R}_t$, and the translation vector $\mathbf{d}_t$.

The branch of rotation subnetwork for learning rotation parameters consists of an LSTM layer, and a fully connected (FC) layer. The parameters related to rotation are obtained with
\begin{equation}
\label{equ:RotationNet}
[\alpha_t,\beta_t,\gamma_t]^{\rm T} = \mathbf{W}_r \mathbf{h}_t^r + \mathbf{b}_r,
\end{equation}
where $\mathbf{h}_t^r \in \mathbb{R}^{N\times 1}$ is the vector of the hidden output of the LSTM layer with $N$ denoting the number of LSTM neurons, $\mathbf{W}_r \in \mathbb{R}^{3\times N}$ and $\mathbf{b}_r\in\mathbb{R}^{3\times 1}$ denote the weight matrix and offset vector of the FC layer, respectively. With the parameters of rotation learned, the rotation matrix $\mathbf{R}_t$ is obtained using (\ref{equ:Rxyz}).

The branch of translation subnetwork for learning translation parameters consists of an LSTM layer, and a FC layer. The translation vector $\mathbf{d}_t$ is calculated as
\begin{equation}
\label{equ:TranslationNet}
\mathbf{d}_t = \mathbf{W}_d \mathbf{h}_t^d + \mathbf{b}_d,
\end{equation}
where $\mathbf{h}_t^d \in \mathbb{R}^{N\times 1}$ is the hidden output vector of its LSTM layer, $\mathbf{W}_d \in \mathbb{R}^{3\times N}$ and $\mathbf{b}_d\in\mathbb{R}^{3\times 1}$ denote the weight matrix and offset vector of the FC layer.

Under the observation viewpoint of the $t^{th}$ frame, the representation of the skeleton $V'_t$ is then obtained through (\ref{equ:transform}). 

\textbf{Main LSTM Network.} The LSTM network is capable of modeling long-term temporal dynamics and automatically learning feature representations. Similar to the designs in \cite{zhu2015co,AAAI17Atte}, we build a main LSTM network by stacking three LSTM layers, followed by one FC layer with a SoftMax classifier. The number of neurons of the FC layer is equal to the number of action classes. 

\textbf{End-to-End Training.} The entire network is end-to-end trainable. We use cross-entropy loss as the training loss \cite{AAAI17Atte}. The gradients of loss flow back not only within each subnetwork, but also from the main LSTM network to the view adaptation subnetwork. Let us denote the loss back-propagated to the output of the view adaptation subnetwork as $\bm{\epsilon}_{v'_{t,j}}\!\in\mathbb{R}^{1\times 3}$, where $j \in (1,\cdots,J)$ and $J$ is the number of joints in a frame. The loss back-propagated to the output of the branch for determining the translation vector of $\mathbf{d}_t$ is 

\begin{equation}
\label{equ:error_dt}
\bm{\epsilon}_{\mathbf{d}_t} = -J  \bm{\epsilon}_{v'_{t,j}} \mathbf{R}_{t},
\end{equation}
Similarly, the loss back-propagated to the output of the branch for determining the rotation parameters can be obtained. For example, the loss back-propagated to the output of $\alpha_t$ is
\begin{equation}
\label{equ:error_dt}
{\epsilon}_{\alpha_t} = \bm{\epsilon}_{v'_{t,j}}  \frac{\partial \mathbf{R}_t}{\partial \alpha_t} \sum_{j=1}^{j=J}(\mathbf{v}_{t,j} - \mathbf{d}_t).
\end{equation}

With the end-to-end training feasible, the view adaptation model is guided to select the suitable observation viewpoints for enhancing the recognition accuracy. 

Our proposed system automatically chooses the suitable observation viewpoints based on the contents, rather than using human predefined criteria, and the view adaptation model is optimized for high accuracy recognition. 

\subsection{View Adaptive Convolution Neural Network (VA-CNN)}
\label{subsec:cnn-va}
	As illustrated by the bottom branch in Fig.~\ref{fig:framework} as marked by VA-CNN, a skeleton sequence is mapped to an image map, referred to as skeleton map, to facilitate the spatio-temporal dyanmics modeling by ConvNet. We use a view adaptation subnetwork constructed by convolution layers and a fully connected layer to learn and determine the sequence-level observation viewpoint, {\it i.e.}, with transform parameters of $\alpha, \beta, \gamma, \mathbf{d}$ (as discussed in section \ref{subsec:formulation} with subscript removed). A main ConvNet follows to explore the spatial and temporal relations and performs the feature extraction from the view-adapted skeleton map for the action recognition, from end to end.
	
	\textbf{Mapping Skeletons to Image.} Similar to that in \cite{du2015skeleton}, we transform a skeleton sequence into an image, with columns representing different frames while rows representing different joints. The 3D coordinate values for $X$, $Y$, and $Z$ are treated as the three channels of an image.

	Considering the differences of values in 3D skeleton and image, similar to \cite{du2015skeleton}, we normalize the pixel values to be within 0-255 by
	\begin{eqnarray}
	\label{equ:normlize}
	\begin{aligned}
	{\bf{u}_{t,j}} = floor(255 \times \frac{{\bf{v}_{t,j}} - \bf{c}_{min}} {c_{max} - c_{min}}),
	\end{aligned}
	\end{eqnarray}
	where $\bf{v}_{t,j}$ denotes the 3D coordinates of the $j^{th}$ joint of the $t^{th}$ frame in a skeleton sequence, $\bf{u}_{t,j}$ denotes the corresponding pixel value of the normalized image map, $c_{max}$ and $c_{min}$ are the maximum and minimum of all the joint coordinates in the training dataset respectively, ${\bf{c}}_{min}=[c_{min},c_{min},c_{min}]^{\rm T}$, $floor$ is the greatest integer function.

	\textbf{View Adaptation Subnetwork.} 
	Observed from a new observation viewpoint, the skeleton representation of the $j^{th}$ joint in the $t^{th}$ frame  $\bf{v}_{t,j}$ is transformed to $\bf{v'}_{t,j}$ according to (\ref{equ:transform}). Correspondingly, the pixel value of the skeleton map under the new observation viewpoint is approximated as
	\begin{eqnarray} 
	{\bf{u'}_{t,j}} \!\!\!& = \!\!\!& 255 \times \frac{{\bf{v'}}_{t,j} - {\bf{c_{min}}}}{ c_{max} - c_{min}} \label{eq:normlize_va} \\ 
	\!\!\!& = \!\!\!& {\bf{R}}_{t,j}  {\bf{u}}_{t,j} +  255 \times \frac{{\bf{R}}_{t,j}  ( {\bf{c}}_{min} \!\!- \!{\bf{d}}_{t,j}) \!\!- \!{\bf{c}}_{min}}{ {c}_{max} - {c}_{min}}. \label{eq:normlize_va_rep} 
	\end{eqnarray} 
	Note that (\ref{eq:normlize_va_rep}) is derived from (\ref{equ:transform}) and (\ref{equ:normlize}).
	
	The CNN-based view adaptation network is designed to learn and determine the observation viewpoint of each skeleton sequence and performs the transform on the skeleton map. We build the view adaptation subnetwork by stacking some convolutional layers and a fully connected layer to regress the transformation parameters, {\it i.e.,} $\alpha, \beta, \gamma$ for ${\bf{R}}_{t,j}$, and $\mathbf{d}_t$.  Based on these parameters and (\ref{eq:normlize_va_rep}), a transform layer  transforms each pixel in the skeleton map to a new representation in the observation viewpoint. Thus, a new skeleton map corresponding to new observation viewpoint is obtained. 
	
	{Note that we have tried regressing frame level parameters $\mathcal{T}_t=\{\alpha_t, \beta_t, \gamma_t, \mathbf{d}_t\}$, which corresponds to $6 \times T$} parameters for a skeleton map with a width of $T$-pixels, and regressing sequence level parameters $\mathcal{T}=\{\alpha, \beta, \gamma, \mathbf{d}\}$, which corresponds to 6 parameters for a skeleton map. Even though the frame level parameters are more flexible and powerful in theory, we find in practice, the design with sequence level parameters provide superior performance for the ConvNets. The reason might be that fewer parameters are easy to learn.
	
	\textbf{Main ConvNet.} With the transformed skeleton map as input, we can use an existing ConvNet, {\it e.g., ResNet \cite{he2016deep} and AlexNet \cite{krizhevsky2012imagenet}}, for classification. 
	
	\textbf{End-to-End Training.} As shown by the bottom branch in Fig. \ref{fig:framework}, similarly, the view adaptation subnetwork, and the main ConvNet is end-to-end trainable, with the view determination optimized by minimizing the classification loss. 	
	
	\subsection{Two Stream Fusion (VA-fusion)}
	\label{subsec:fusion}
	Similar to the fusion strategy in \cite{wang2016temporal}, we can take a weighted fusion method to combine the scores of the VA-RNN stream and VA-CNN stream to obtain the final score. Considering the performance gap between the two streams, we give more credit to the VA-CNN stream by setting its weight to 4 and that of the VA-RNN stream to 1, which is experimentally determined. We have tried learning the weights but it does not outperform the simple fusion approach. 
	
	Note that we do not investigate more complicated fusion scheme since that is not the focus of this work. We try to allow each stream to be as independent as possible, which can facilitate the design choices in practice. The users can choose the scheme which better meets the requirements, such as speed, hardware, and storage among VA-RNN, VA-CNN, and VA-fusion.   

	\subsection{Model Implementation and Training}
	\label{subsec:modelimp}
	\textbf{Model Architecture.} For VA-RNN, we build our model using recurrent neural networks with LSTM. We use 100 neurons for each LSTM layer. For the main LSTM network, we stack three LSTM layers. For the view adaptation subnetworks, we only use one LSTM layer followed by one fully connected layer to learn the transformation parameters. 
	
	For VA-CNN, we build our model using convolutional neural networks. For our main ConvNet, similar to \cite{du2015skeleton, li2017skeleton}, we use ResNet-50 with pretrained parameters from ImageNet for classification. The view adaptation subnetwork is constructed by stacking two convolutional layers and one fully connected layer. Note that the two convolutional layers are both followed by a batch normalization layer and a Relu activation layer. A max-pooling layer is used to reduce the resolution after the second convolutional layer and finally a fully connected (FC) layer is used to regress the parameters related to view transformation. We set the number of kernels to 128 for all these convolution layers. For each convolutional layer, we set the kernel size to 5 and stride to 2.

	\textbf{View Enriching by Data Argumentation.} All the data have a limited range of capturing viewpoints. This is common especially in practical application scenarios. To enhance the ``power" of our view adaptation model, at the sequence level, we perform view enriching by rotating the skeleton around the $X$, $Y$, and $Z$ axes by some degrees during training procedure. This is expected to alleviate the overfitting problem especially on small datasets, and reinforce the view adaptation model.
	
	\textbf{Model Training.} We train each stream of view adaptive neural network by minimizing the cross-entropy loss end-to-end. Similar to that in \cite{wang2016temporal}, we fuse the two streams by weighted averaging the scores and obtain the classification probability after the SoftMax function.  	
	
\section{Experimental Results}
\label{sec:results}

We evaluate the effectiveness of our proposed view adaptation frameworks on five benchmark datasets, including the NTU RGB+D dataset \cite{Shahroudy_2016_CVPR}, the SYSU Human-Object Interaction dataset \cite{hu2015jointly}, the UWA3D dataset \cite{rahmani2016histogram}, the Northwestern-UCLA dataset \cite{wang2014cross}, and the SBU Kinect Interaction dataset \cite{yun2012two}.

In the following, we first describe the datasets and our experiment settings in Section \ref{subsec:datasets}.
For the ablation studies in Section \ref{subsec:ablation}, the effectiveness of the proposed view adaptation model is analyzed and demonstrated. We also make comparison with other view-invariant strategies. The influence of  parameters are analyzed. To further understand the view adaptation model, we perform analysis through visualization and some failure cases are discussed. Section \ref{subsec:comparison} presents the performance comparisons with the state-of-the-art approaches on the five datasets respectively. The results demonstrate that our scheme consistently achieves the best performance on all datasets. Some comparative analyses of VA-RNN and VA-CNN are presented in Section \ref{subsec:complexity}.

\subsection{Datasets and Experimental Settings} 
\label{subsec:datasets}

\textbf{Datasets.} \emph{NTU RGB+D Dataset (NTU) \cite{Shahroudy_2016_CVPR}.} This Kinect captured dataset is currently the largest dataset with RGB+D videos and skeleton data for human action recognition, with 56880 video samples. It contains 60 different action classes including daily, mutual, and health-related actions. Each subject has 25 joints. The various setups of cameras, capturing views, and different facing orientations of the subjects, result in a great diversity of sample viewpoints. There are two standard evaluations, {\it{i.e.}}, Cross-Subject (CS), where the 40 subjects are split into the training and testing groups, and Cross-View (CV), where the samples of cameras 2 and 3 are used for training and those of camera 1 for testing. It is a challenging dataset for action recognition because of the large amount of videos, various subjects, and the difference of camera views.

\emph{SYSU 3D Human-Object Interaction Set (SYSU) \cite{hu2015jointly}.} This Kinect captured dataset contains 12 actions performed by 40 subjects. It has 480 sequences in total. Each subject has 20 joints. This dataset is challenging for high similarity among activities.

\emph{UWA3D Multiview Activity II Dataset (UWA3D) \cite{rahmani2016histogram}.} This Kinect captured dataset contains 30 actions performed by 10 different subjects. The videos are captured from 4 different views: front view (V1), left side view (V2), right side view (V3), and top view (V4). It has 1075 sequences in total. This dataset is challenging because of the diversity of viewpoints, self-occlusion, and high similarity among activities.

\emph{Northwestern-UCLA dataset (N-UCLA) \cite{wang2014cross}.} This Kinect captured dataset contains 1494 videos of 10 actions. These actions are performed by 10 subjects, repeated 1 to 6 times. There are three views. Each subject has 20 joints.

\emph{SBU Kinect Interaction Dataset (SBU) \cite{yun2012two}.} This Kinect captured dataset is an interaction dataset with each action performed by two subjects. It contains 282 sequences of 8 classes. Each subject has 15 joints. 

Note that  both the SYSU and SBU dataset are captured by a single camera with one primary viewpoint. However, different subjects may perform actions at different locations with different distances to the camera and orientations.  

\textbf{Experimental Settings.} For VA-RNN, We set the batch size to 256 for the NTU dataset and 32 for other datasets in considering the small sizes of the other datasets. For the view adaptation subnetwork, we initialize the fully connected layer parameters to zeros for efficient training. Dropout \cite{srivastava2014dropout} with a probability of 0.5 is used to alleviate overfitting. Gradient clipping similar to \cite{sutskever2014sequence} is used by enforcing a hard constraint on the norm of the gradient (not exceeding 1) to avoid the gradient explosion problem. Adam \cite{kingma2014adam} is adapted to train all networks, and the initial learning rate is set to 0.005 for all datasets. 

For VA-CNN, we set the batch size to 32. For the view adaptation subnetwork, we initialize the fully connected layer parameters to zeros for efficient training. Adam \cite{kingma2014adam} is adapted to train all networks, and the initial learning rate is set to 0.0001 for all datasets. Skeleton maps are resized to 224$\times$224.

In the UWA3D and N-UCLA datasets, at the sequence level, we rotate the skeleton around the $X$, $Y$, and $Z$ axes by some degrees which are generated randomly from -90 to 90 during training procedure. Considering that the ranges of view variation on other datasets are smaller, at the sequence level, we rotate the skeleton around the $X$, $Y$, and $Z$ axes by some degrees which are generated randomly from -17 to 17. 

\begin{table}[t] 
	\centering
	\caption{Comparisons of pre-processing methods and our view adaptation model on the NTU dataset.}
	\label{table:In-component}%
	\begin{tabular}{c|c|c|c}
		\hline
		& Methods & CS & CV \\
		\hline
		wo/ pre-proc. & Raw + RNN & 66.3 & 73.4\\
		\hline
		\multirow{8}[3]{*}{Pre-proc.} & \emph{S-trans + RNN} & 76.0 & 82.3\\
		\cline{2-4}       & F-trans + RNN & 75.1 & 80.5 \\
		\cline{2-4}       & \tabincell{c}{S-trans$\&$S-rota(w.r.t. shoulder)\\ + RNN} & 75.8  & 85.1  \\
		\cline{2-4}       & S-trans$\&$S-rota + RNN & 76.4 & 85.4 \\
		\cline{2-4}       & \tabincell{c}{S-trans$\&$F-rota(w.r.t. shoulder) \\+ RNN }& 75.8 & 84.9  \\
		\cline{2-4}       & S-trans$\&$F-rota + RNN & 75.0 & 85.1 \\
		\cline{2-4}       & F-trans$\&$F-rota + RNN & 74.1 & 83.9 \\
		\hline
		\multirow{3}[3]{*}[1.5mm]{View adap.} & VA-trans + RNN & 77.7 & 84.9 \\
		\cline{2-4}       & VA-rota + RNN & 79.4 & 87.1 \\
		\cline{2-4}       & \textbf{VA-RNN} & \textbf{79.4} & \textbf{87.6} \\
		\hline
	\end{tabular}%
\end{table}%

	\begin{table*}[t]
	\centering
	\caption{Effectiveness (in accuracy(\%)) of data augmentation on \emph{S-trans} and \emph{VA} schemes.}
	\label{tab:augmentation}
	\begin{tabular}{c|c|c|c|c|c|c|c|c}
		\hline
		\multicolumn{2}{c|}{\multirow{2}{*}{Datasets}} & \multicolumn{2}{c|}{NTU}     & \multicolumn{2}{c|}{SYSU}     & \multirow{2}{*}{UWA3D} & \multirow{2}{*}{N-UCLA}   & \multirow{2}{*}{SBU} \\ \cline{3-6}
		\multicolumn{2}{c|}{}                          & CS            & CV            & setting-1     & setting-2     &                        &                          &                      \\ \hline
		\multirow{4}{*}{RNN-based}      & S-trans+RNN        & 76.0          & 82.3          & 76.3          & 75.6          & 57.4                   & 67.4                                 & 93.8                 \\ 
		\cline{2-9} 
		& S-trans+RNN(aug.)    & \textbf{77.0}          & \textbf{85.0}          & \textbf{79.3 }         & \textbf{78.5}          & \textbf{69.5    }               & \textbf{80.9    }                      & \textbf{93.9}               \\ \cline{2-9} 		
		\cline{2-9} 
		& VA-RNN              & 79.4          & 87.6          & 77.5          & 76.9          & 58.3                   & 70.7                 & 95.9                 \\  
		\cline{2-9} 
		& VA-RNN(aug.)          & \textbf{79.8} & \textbf{88.9} & \textbf{80.5} & \textbf{79.8} & \textbf{73.6}          & \textbf{84.1}                & \textbf{97.5}        \\ \hline
		\multirow{4}{*}{CNN-based}       & S-trans+CNN        & 87.5          & 92.2          & 82.1          & 80.9          & 67.2                & 73.9                                 & 86.7                 \\ 
		\cline{2-9} 
		& S-trans+CNN(aug.)    & \textbf{87.9}          & \textbf{93.5}      & \textbf{84.2}        &\textbf{ 83.4}          & \textbf{77.0 }                  & \textbf{85.7}                        & \textbf{93.0 }                \\ \cline{2-9} 
		& VA-CNN              & 88.2          & 93.8          & 83.6          & 82.9          & 71.2                   & 81.7                    & 90.3                 \\ 
		\cline{2-9} 
		
		& VA-CNN(aug.)         & \textbf{88.7} & \textbf{94.3} & \textbf{85.1} & \textbf{84.8} & \textbf{79.3}          & \textbf{86.6}                & \textbf{95.7}        \\ \hline
	\end{tabular}
\end{table*}

\subsection{Ablation Study}
\label{subsec:ablation}

\subsubsection{Comparison with Other Pre-processing Strategies}
\label{subsec:pre-processing}

Some approaches use human defined criteria to pre-process the skeletons to reduce the challenges caused by view variations \cite{CVPR15HRNN,zhu2015co, Shahroudy_2016_CVPR, liu2016spatio,AAAI17Atte,liu2017global,li2017RNNTree}. We make comparison on the effectiveness of those strategies and our view adaptation model. Considering that the NTU RGB+D dataset is currently the largest dataset and is representative, we perform our in-depth analyses on this dataset under the framework of recurrent neural networks and show the results in Table \ref{table:In-component}.  

\emph{VA-RNN} is our proposed view adaptation scheme under the RNN framework, which automatically changes the observation viewpoints in the network. In comparison, \emph{S-trans+RNN} is our baseline scheme without enabling the view adaptation model, {\it i.e.}, the switch $s_{rota}$ and $s_{trans}$ are both off, {\it i.e.} $V'_t = V_t$. Note that the input $V_t$ is the same as that of our view adaptation schemes, where the global coordinate system is moved to the body center of the first frame for the entire sequence to be insensitive to the initial position (see section \ref{subsec:formulation}). We refer to this pre-processing as sequence level translation, {\it i.e.}, \emph{S-trans}.

From Table \ref{table:In-component}, we observe that the proposed view adaptation scheme outperforms the \emph{S-trans+RNN} by 3.4\% and 5.3\% in accuracy for the CS and CV settings, respectively.  We notice that rotation-only adaptation \emph{VA-rota+RNN} seems to be more effective than translation-only adaptation \emph{VA-trans+RNN}. That is because most actions in this dataset are performed without shifting positions during the occurrence.

One may wonder about the performance when using the pre-processed skeletons, based on the widely used human defined processing criteria, before inputing them to the main RNN Network. Such pre-processings follow human defined rules to determine the viewpoints. We denote the pre-processing based schemes as \emph{C+RNN}, where \emph{C} indicates the pre-processing strategy, {\it e.g.}, \emph{F-trans+RNN}. The $3^{rd}$ to  $9^{th}$ rows show the performance of schemes using different pre-processing strategies. \emph{F-trans} means performing frame level translation to have the body center at the coordinate system origin for each frame. \emph{S-rota} is the sequence level rotation with the rotation parameters calculated from the first frame, which is to fix the $X$-axis to be parallel to the vector from ``left shoulder" to ``right shoulder", $Y$-axis to be parallel to the vector from ``spline base" to ``spine", and $Z$-axis corresponding to the new $X\!\times\!Y$. Similarly, \emph{F-rota} is the frame wise rotation. {For \emph{S/F-rota(w.r.t. shoulder)}, only the rotation pre-processing to make the X-axis to be parallel to the vector from ``left shoulder" to ``right shoulder" is performed at sequence/frame level.} \emph{F-trans}$\&$\emph{F-rota} means both \emph{F-trans} and \emph{F-rota} are performed, which is similar to the pre-processing in \cite{Shahroudy_2016_CVPR, liu2016spatio,AAAI17Atte}. The scheme \emph{Raw+RNN}  in the $2^{nd}$ row denotes a scheme which uses the original skeleton without any pre-processing as the input to the Main RNN Network. Note that for 3D skeletons, the distance of a subject to the camera does not influence the scale of the skeletons. Therefore, the scaling operation is not considered in our framework.
\begin{savenotes}
	\begin{table*}[t]
		\centering
		\caption{Accuracy (\%) comparisons of two types of powerful baseline schemes, {\it{i.e.}}, the schemes with sequence translation pre-processing strategy (\emph{S-trans+RNN}(aug.) and \emph{S-trans+CNN}(aug.)), the schemes with sequence translation and rotation pre-processing strategy (\emph{S-trans}\&\emph{S-rota+RNN} and \emph{S-trans}\&\emph{S-rota+CNN}), and our schemes with view adaptation. Note for the two types of baseline schemes, we mark the better one by underline.}
		\label{table:all-basline-ours}
		\begin{tabular}{c|c|c|c|c|c|c|c|c}
			\hline
			\multicolumn{2}{c|}{\multirow{2}{*}{Datasets}} & \multicolumn{2}{c|}{NTU}     & \multicolumn{2}{c|}{SYSU}     & \multirow{2}{*}{UWA3D} & \multirow{2}{*}{N-UCLA}   & \multirow{2}{*}{SBU} \\ \cline{3-6}
			\multicolumn{2}{c|}{}                          & CS            & CV            & setting-1     & setting-2     &                        &                          &                      \\ \hline
			\multirow{3}{*}{RNN-based} 
			& S-trans\&S-rota+RNN   & 76.4 & \underline{85.4}  & 76.1 & 75.6 &\underline{70.3} & 80.5 & \underline{95.5} \\
			\cline{2-9} 
			& S-trans+RNN(aug.)   & \underline{77.0}  & 85.0 & \underline{79.3}   & \underline{78.5}  &{69.5} & \underline{80.9} & 93.9                 \\ \cline{2-9} 
			
			& VA-RNN(aug.)  & \textbf{79.8} & \textbf{88.9} & \textbf{80.5} & \textbf{79.8} & \textbf{73.6} & \textbf{84.1}  & \textbf{97.5}        \\ \hline
			
			\multirow{3}{*}{CNN-based}   
			& S-trans\&S-rota+CNN  & 87.4  & 93.3   & 81.0 & 80.3 & 75.7  & 83.2  & 87.9                \\ 
			\cline{2-9} 
			& S-trans+CNN(aug.)  & \underline{87.9}  &\underline{93.5}  & \underline{84.2}  & \underline{83.4}  & \underline{77.0}   & \underline{85.7} & \underline{93.0}                 \\ \cline{2-9} 
			& VA-CNN(aug.)         & \textbf{88.7} & \textbf{94.3} & \textbf{85.1} & \textbf{84.8} & \textbf{79.3}          & \textbf{86.6}                & \textbf{95.7}        \\ \hline
		\end{tabular}
	\end{table*}
\end{savenotes}

From Table \ref{table:In-component}, we have the following observations and conclusions. (1) Our final scheme significantly outperforms the commonly used pre-processing strategies. In comparison with \emph{F-trans}$\&$\emph{F-rota}+\emph{RNN} \cite{Shahroudy_2016_CVPR, liu2016spatio,AAAI17Atte}, our scheme achieves an improvement of 5.3\% and 3.7\% in accuracy for the CS and CV settings, respectively. In comparison with \emph{S-trans}$\&$\emph{S-rota}+\emph{RNN}, our scheme achieves an improvement of 3.0\% and 2.2\% in accuracy. (2) Frame wise pre-processing is inferior to the sequence level pre-processing, because the former loses more information, {\it e.g.}, the motion across frames. (3) Being insensitive to the initial position of an action, \emph{S-trans+RNN} significantly outperforms \emph{Raw+RNN}, the scheme with raw skeletons as input. (4) {Some human-defined pre-processings, such as \emph{S-trans}\&\emph{S-rota}, \emph{S-trans}\&\emph{F-rota}, \emph{S-trans}\&\emph{S-rota(w.r.t. shoulder)} and \emph{S-trans}\&\emph{F-rota(w.r.t. shoulder)} achieve superior results to those without rotation processing on the CV setting. Because such pre-processings can reduce the diversity of viewpoints and alleviate the viewpoint mismatch problem between the training and testing samples.}

\subsubsection{Influence of Data Augmentation}
\label{subsec:augmentation}

Data augmentation by increasing the viewpoint diversity during the training procedure can alleviate the viewpoint mismatch problem between training and testing sets. It strengthens the capability of both baseline scheme S-trans and our VA scheme. The VA module benefits from data augmentation by ``seeing" more views during training to learn how to make transformation for various views. We will show the effectiveness of data augmentation on both baseline scheme \emph{S-trans} and the proposed view adaptation model \emph{VA} in Table \ref{tab:augmentation}. \emph{S-trans+RNN/CNN} and \emph{VA-RNN/CNN} denote the baseline schemes and the proposed view adaptation schemes without data augmentation, respectively. \emph{S-trans+RNN/CNN}(aug.) and \emph{VA+RNN/CNN}(aug.) denote the schemes with data augmentation(aug.).

We can see that, for the CNN-based networks, data augmentation improves the performance by 0.4\% and 1.3\% on the CS and CV settings of NTU dataset, respectively. For the CV setting, the viewpoints of the testing data are different from those of the training data. Thus, data augmentation by increasing the viewpoints can bring larger gain on the CV setting, which makes some unseen testing views seen during the training process. For the UWA3D and N-UCLA datasets which are under CV setting, the view differences are large where there is even a top view while the other views are captured by cameras located at nearly horizontal positions. Data augmentation enables the training process to ``see" the testing views and thus it even brings gains of 9.8-11.8\%. Data augmentation mainly addresses the mismatch between training viewpoints and testing viewpoints by increasing the diversity of training viewpoints.

In addition, with the help of data augmentation, \emph{VA-RNN}(aug.) and \emph{VA-CNN}(aug.) improve the performance significantly in comparison with \emph{VA-RNN} and \emph{VA-CNN}, especially for the UWA3D and N-UCLA datasets. For the UWA3D and N-UCLA datasets, data augmentation brings gains of 8.1\% and 4.9\% on VA-CNN. The main reason is that the \emph{VA-RNN} and \emph{VA-CNN} models are not able to transform the skeleton sequence of testing set to good learned view when the view of training and testing sets mismatch with each other significantly. However, thanks to the data augmentation, the \emph{VA-RNN}(aug.) and \emph{VA-CNN}(aug.) models could ``see" plenty of views during training and be capable of transforming skeletons of both training and testing sets to suitable learned view.

Data augmentation is an efficient and necessary technique for both the baseline and the proposed view adaptation schemes. Hereafter, our experiments are all conducted with data augmentation.

\subsubsection{Effectiveness of the View Adaptation Model}
\label{subsec:efficiency}

We will compare our proposed \emph{VA} model with the two powerful baselines in Table \ref{table:all-basline-ours}. \emph{S-trans}\&\emph{S-rota+RNN} and \emph{S-trans}\&\emph{S-rota+CNN} are the baseline schemes with both human defined translation and rotation pre-processing. Note that for \emph{S-trans}\&\emph{S-rota}, data augmentation by rotating the skeleton sequence to increase the view diversity should not be performed since the purpose of the rotation pre-processing is to align the viewpoints. \emph{S-trans+RNN}(aug.) and \emph{S-trans+CNN}(aug.) are another type of baseline schemes with only translation pre-processing, and data augmentation is performed.

\textbf{\emph{View consistence by view adaptation versus pre-processing.}} From Table \ref{table:all-basline-ours}, we find that, in comparison with the human defined rotation pre-processing, the view adaptation model consistently achieves superior performance. The human defined pre-processing strategy is not optimized for recognition performance. Human body is non-rigid and the definition of rotation criteria is not always suitable for the purpose of orientation alignment. Our scheme leverages the network to automatically determine the suitable viewpoints, trained by optimizing the classification accuracy.

\textbf{\emph{View adaptation versus data augmentation.}} From Table \ref{table:all-basline-ours}, \emph{VA-RNN}(aug.) and \emph{VA-CNN}(aug.) outperform \emph{S-trans+RNN}(aug.) and \emph{S-trans+CNN}(aug.) for all datasets. Viewpoint is a distractor rather than a discriminative characteristic for action recognition. Therefore, for a network, when there is no mismatch between training and testing viewpoints ({\it{e.g.}}, after data augmentation), it should be more challenging to deal with diverse viewpoints than only dealing with consistent viewpoints. Our view adaptation scheme \emph{VA-RNN} intends to transform the diverse viewpoint to a consistent viewpoint to alleviate the difficulty. The learned consistent viewpoint is beneficial to learning discriminative features.

In conclusion, our proposed \emph{VA-RNN}(aug.) and \emph{VA-CNN}(aug.) consistently achieve the best performance in comparison with two powerful baseline schemes for all datasets.

\begin{table}[t]
	\centering
	\caption{Accuracy (\%) comparisons on different numbers of LSTM layers for the main LSTM network (S-trans+RNN(aug.)), and different numbers of convolutional layers for the main ConvNet (S-trans+CNN(aug.)) on the NTU dataset.}
	\label{table:layers}
	\resizebox{0.49\textwidth}{!}{
	\begin{tabular}{c|c|c|c|c}
		\hline
		Main Network & Structure & \#Param.(M) & CS    & CV \\
		\hline
		\multirow{3}[12]{*}{S-trans+RNN(aug.)} & 1 LSTM layer & 0.11  & 74.5  & 82.0  \\
		\cline{2-5}          & 2 LSTM layers & 0.19  & 76.2  & 84.7  \\
		\cline{2-5}          & 3 LSTM layers & 0.27  & 77.0  & 85.0  \\
		\cline{2-5}          & 4 LSTM layers & 0.35  & 76.9  & 83.9  \\
		\cline{2-5}          & 5 LSTM layers & 0.43  & 76.2  & 84.2  \\
		\cline{2-5}          & 6 LSTM layers & 0.51  & 76.6  & 84.4  \\
		\hline
		{VA-RNN(aug.)} & 3+2 LSTM layers & 0.47  & 79.8  & 88.9  \\
		\hline
		\hline
		\multirow{2}[8]{*}{S-trans+CNN(aug.)} & ResNet18 & 11.21 & 86.5  & 93.1  \\
		\cline{2-5}          & ResNet50 & 23.63 & 87.9  & 93.5   \\
		\cline{2-5}          & ResNet101 & 42.62 & 87.8  & 93.5  \\
		\cline{2-5}          & ResNet152 & 58.27 & 88.2  & 93.4  \\
		\hline
		{VA-CNN(aug.)} & ResNet50+3 layers   & 24.09 & 88.7  & 94.3  \\
		\hline
	\end{tabular}
}
\end{table}%

\begin{figure}[t] 
	\centering
	\begin{subfigure}[t]{0.7\linewidth}
		\centering\includegraphics[width=\textwidth]{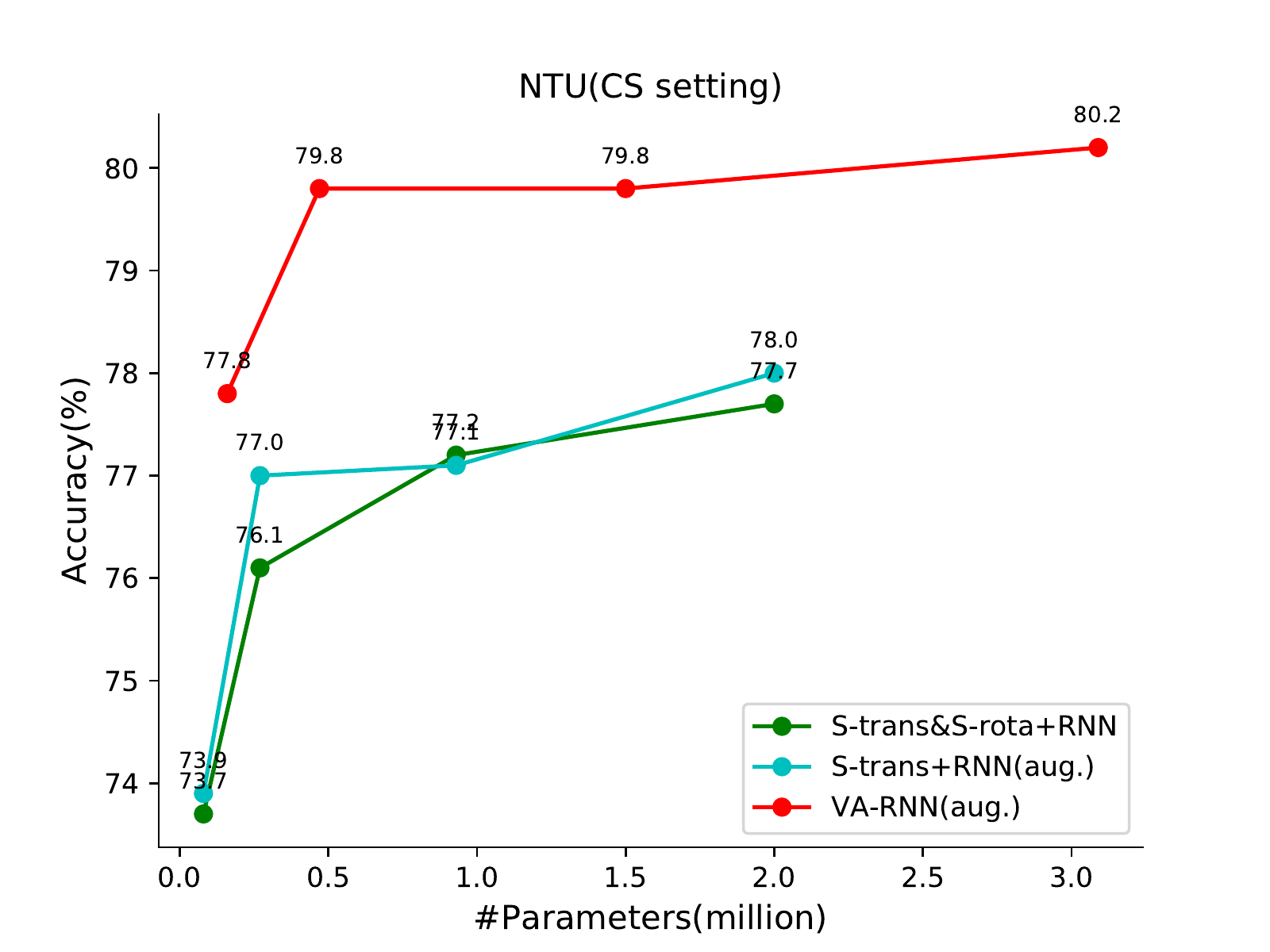}
		\caption{NTU-CS}
		\label{subfig:CS}
	\end{subfigure}	
	\begin{subfigure}[t]{0.7\linewidth}
		\centering\includegraphics[width=\textwidth]{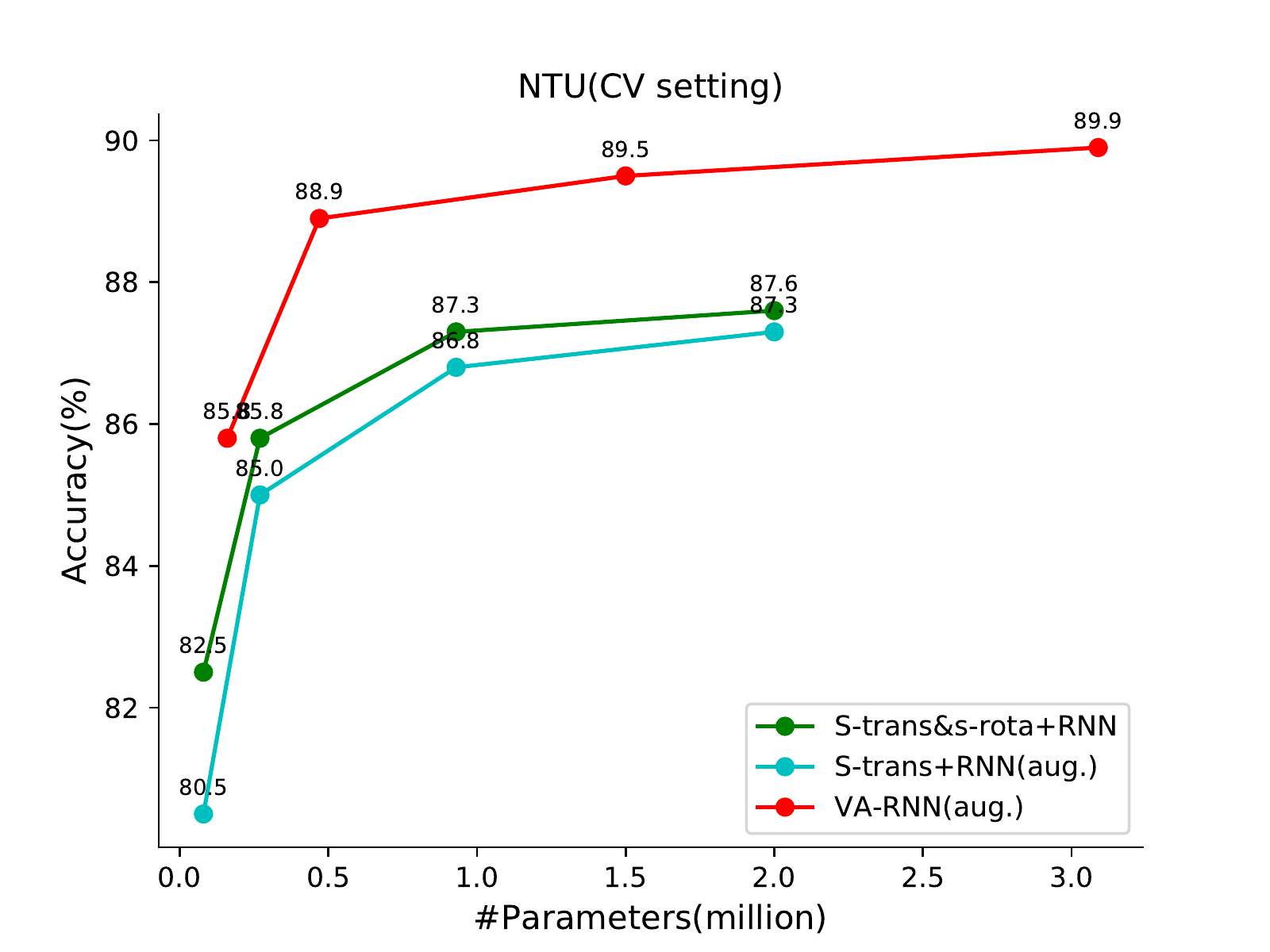}
		\caption{NTU-CV}			
		\label{subfig:CV}
	\end{subfigure}
	\caption{Performance curve for both baselines and the proposed view adaptation schemes based on RNN on the NTU dataset. The horizontal axis denotes the model size, {\it{i.e.}} number of parameters, while the vertical axis indicates the recognition accuracy (\%).}\label{fig:neurons}
\end{figure}

\begin{figure*}[http]
	\centering
	\begin{subfigure}[t]{0.45\linewidth}
		\centering\includegraphics[width=\textwidth]{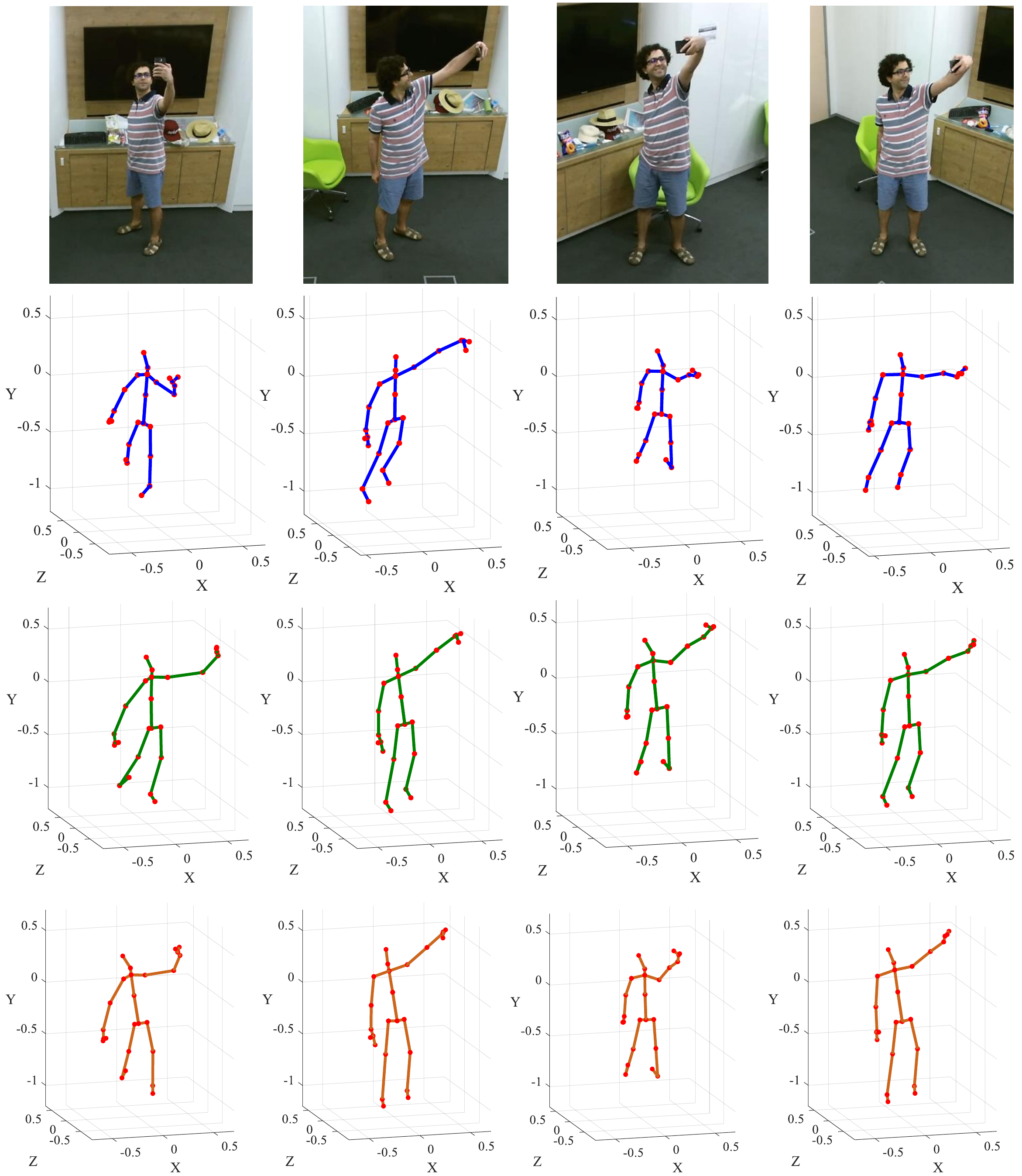}
		\caption{}
		\label{subfig:selfie-VA}
	\end{subfigure}	
	~
	\begin{subfigure}[t]{0.45\linewidth}
		\centering\includegraphics[width=\textwidth]{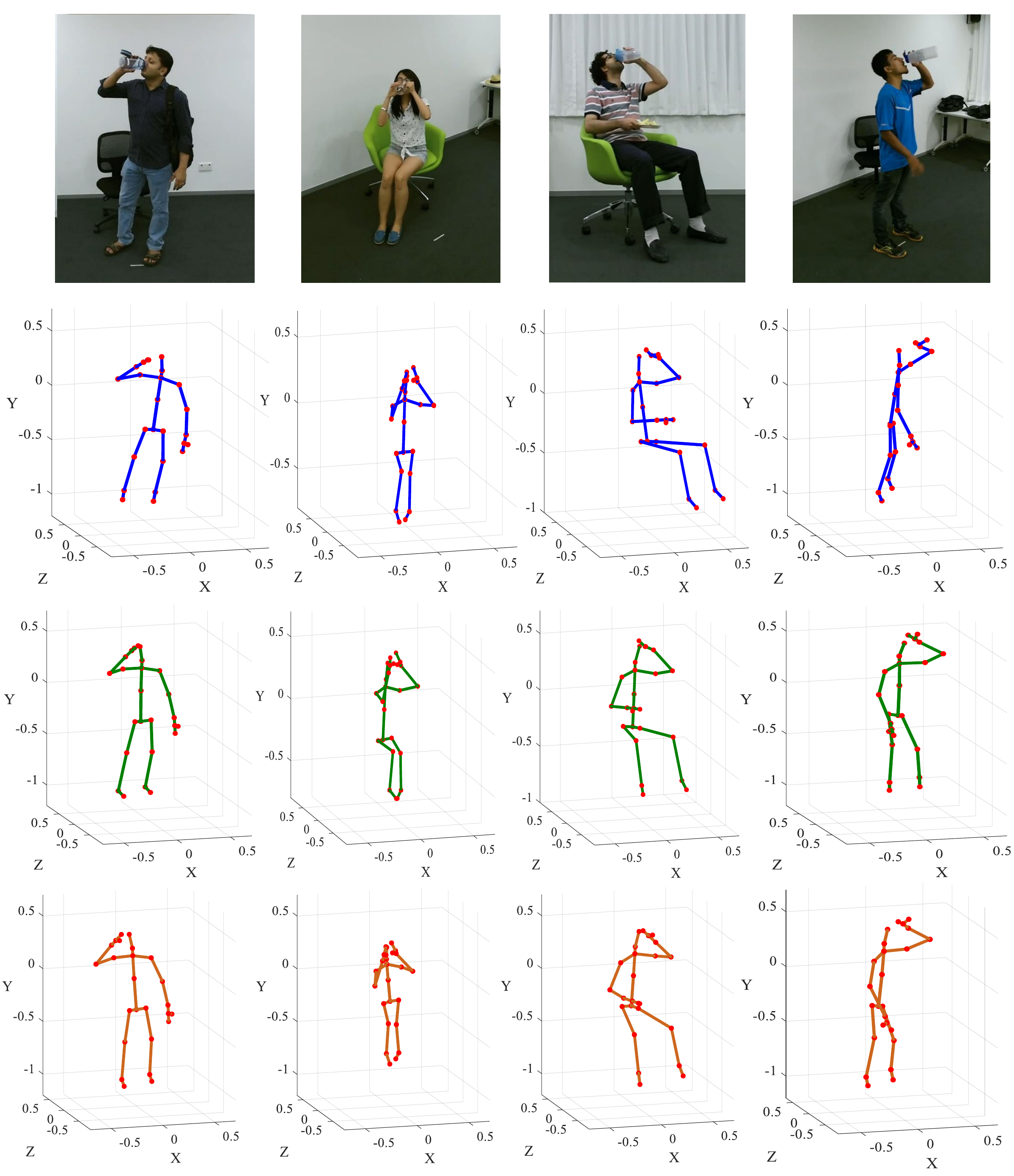}
		\caption{}			
		\label{subfig:Results-actors-VA}
	\end{subfigure}
	\vspace{-2mm}
	\caption[]{Frames of (a) the similar posture captured from different viewpoints for the same subject, and (b) the same action ``drinking" captured from different viewpoints for different subjects. $2^{nd}$ row: original skeletons. $3^{rd}$ row: skeleton representations from the observation viewpoints of our VA-RNN model. $4^{th}$ row: skeleton representations from the observation viewpoints of our VA-CNN model.} \label{fig:action-view}
\end{figure*}

\subsubsection{Influence of Network Parameters}
\label{subsec:neuronnumber}

\emph{VA} module is the subnetwork of the recognition network. Therefore, \emph{VA-RNN} and \emph{VA-CNN} have more parameters than the corresponding base networks. One may wonder whether the gains are brought by the increased number of parameters or the proposed view adaptation modules. There are two ways to increase the model size of a network: (1) Stacking more RNN or CNN layers; (2) Increasing the number of LSTM neurons for RNN-based network or convolutional kernels for CNN-based network. Note that we use the \emph{ResNet} as our CNN-based backbone network with pre-trained parameters from ImageNet. Therefore, we have not changed the number of convolutional kernels. Instead, we show the results when changing the number of LSTM neurons of each RNN layer for RNN-based networks. We will discuss these two ways as follows.

\textbf{\emph{Stacking more layers.}} Table \ref{table:layers} shows comparisons between our proposed view adaptation models and the corresponding main classification networks with different number of layers. Each LSTM layers contains 100 neurons. For the RNN-based networks, as the increase of the number of LSTM layers, the performance initially increases but drops after 3 layers. Stacking LSTM layers simply would not achieve better performance significantly. However, our proposed \emph{VA-RNN}(aug.), which includes 5 LSTM layers (3 LSTM layers for the main network and 2 LSTM layers for the VA subnetwork), outperforms the baseline scheme with 5 or 6 LSTM layers by around 3.5\% and 4.5\% for the CS and CV settings, respectively. For the CNN-based networks, we use \emph{ResNet} of different layers as our backbone networks and find that a deeper network does not bring obvious gain. In comparison, our scheme with 53 layers outperforms the baseline scheme with 152 layers by 0.5\% and 0.9\% for the CS and CV settings, respectively.

\textbf{\emph{Increasing the number of LSTM neurons.}} Fig.~\ref{fig:neurons} shows comparisons between our proposed view adaptation models and the corresponding main classification networks for RNN-based networks with different number of LSTM neurons. The baseline models contains 3 LSTM layers and the \emph{VA} model contains 5 LSTM layers. We set the number of neurons of the three models, which include \emph{VA-RNN}(aug.), \emph{S-trans+RNN}(aug.) and \emph{S-trans}\&\emph{S-rota+RNN}, to 50, 100, 200 and 300, respectively. For each model, the number of parameters and the performance increase when we use more neurons. \emph{VA-RNN}(aug.) always outperforms \emph{S-trans+RNN}(aug.) and \emph{S-trans}\&\emph{S-rota+RNN} with similar or fewer parameters for both the CS and CV settings. Note that recognition accuracy increases since more neurons have a higher capability of modeling the evolution of action dynamics for all three schemes, we use 100 neurons for each LSTM layer by default to balance performance, speed and complexity.

In conclusion, increasing parameters simply by stacking more layers or using more number of neurons is not as efficient as our proposed view adaptation module. With similar amount of parameters, our models outperform the baseline models.

\subsubsection{Visualization {and Analysis} from the Learned Views}
\label{subsec:visualization}

The view adaptation subnetworks determine the observation viewpoints (by re-positioning the virtual movable camera) and then transform the input skeleton $V_t$ to the representation $V'_t$ under the new viewpoint for optimizing the recognition performance. We visualize the representations $V_t$ and $V'_t$ for better understanding our models.

Fig.~\ref{fig:action-view} shows the skeletons from different sequences captured from different viewpoints of (a) the similar posture or (b) the same action. The $2^{nd}$ row shows the original skeletons of diverse viewpoints. The $3^{rd}$ row shows the transformed skeletons from our \emph{VA-RNN} model. The transformed skeletons have much more consistent viewpoints even for different subjects and actions. The $4^{th}$ row shows the transformed skeletons from our \emph{VA-CNN} model. Abundant observations on other sequences show that both the \emph{VA-RNN} and \emph{VA-CNN} models are capable of transforming skeletons to much more consistent viewpoints. Note that the consistency of viewpoints after our model is the key factor for the success of our scheme. In addition, we have observed the transformed skeletons on many sequences and found the continuity of an action can be maintained by our models.
 
The view adaptation model transforms the viewpoint of skeleton sequence based on its contents. One may wonder how many frames the system needs to learn well of the transformation parameters \emph{VA-RNN}. We have {looked into} abundant transformed sequences and transformation parameters and have the following observations. First, the network starts to modify the skeleton immediately when receiving the first skeleton frame. However, the learned views of the first several frames are not very good. For the first several frames, the LSTM network has not ``seen" enough information to have a good guess of the views. Second, it takes around 5 to 20 frames to transform the skeleton to a relative stable view.

\subsubsection{Failure Case Discussion}
\label{subsec:failure}
\begin{figure}[t] 
	\begin{center}
		\includegraphics[width=1\linewidth]{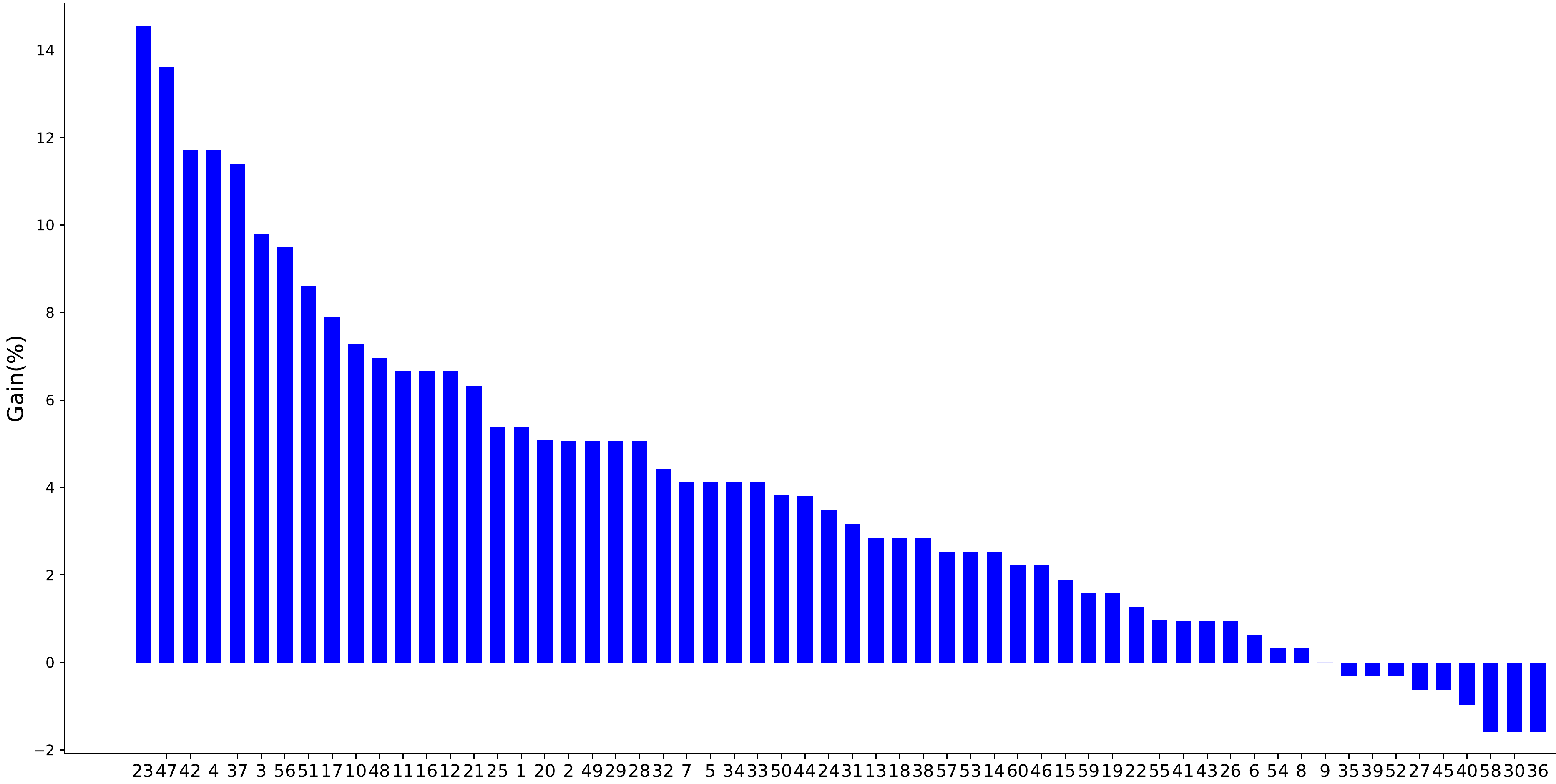}
	\end{center}
	\caption{Performance gain in terms of accuracy (\%) of VA-RNN model with respect to the S-trans+RNN on the NTU dataset for the CV setting. The index of the horizontal axis denotes the Id of action as provided in \cite{Shahroudy_2016_CVPR}. For example, ``23" denotes the action \emph{hand waving}.}
	\label{fig:hist_class}
\end{figure}

One may wonder about the failure cases of the \emph{VA-RNN} or \emph{VA-CNN} models in comparison with the baseline schemes \emph{S-trans+RNN} or \emph{S-trans+CNN}. We have made abundant observations and found even for the misclassified samples, the skeletons are still transformed to consistent views by our proposed models. Fig.~\ref{fig:hist_class} shows the histogram of the performance gains of our \emph{VA-RNN} model with respect to the baseline model \emph{S-trans+RNN} on the NTU dataset for the CV setting. The gain values are obtained by subtracting the accuracy of the \emph{S-trans-RNN} from that of the \emph{VA-RNN} on each action classes. We can see that, for most classes, our scheme outperforms the baseline scheme. For the classes on which performance drops, we do not find special reasons and this can be considered as noise. As we know, even for a same network, there is performance fluctuation when different initialization seeds are utilized. Similar phenomenon can be observed for \emph{VA-CNN} scheme and we do not show to save space.

\subsection{Comparisons to Other State-of-the-Art Approaches}
\label{subsec:comparison}
In this section, we show the performance comparisons between our proposed two stream view adaptation scheme \emph{VA-fusion(aug.)} and other state-of-the-art approaches on these datasets. The performance of  \emph{VA-RNN(aug.)} and \emph{VA-CNN(aug.)} are also presented.

\subsubsection{NTU Dataset}
We follow the standard CS and CV protocols proposed by \cite{Shahroudy_2016_CVPR} to evaluate the performance. 	
We compare our method with the deep learning based approaches that also leverage RNN or CNN with skeleton data \cite{CVPR15HRNN,liu2016spatio,Shahroudy_2016_CVPR,AAAI17Atte,liu2017global,li2017RNNTree,liu2017enhanced}, and some traditional approaches using handcrafted features \cite{evangelidis2014skeletal,vemulapalli2014human}. The results are shown in Table \ref{table:NTU}.  This dataset contains as many as 80 viewpoints which will bring difficulty to action recognition. Thanks to the view adaptation models which make the views consistent,  \emph{VA-RNN(aug.)} and \emph{VA-CNN(aug.)} outperform the baseline schemes significantly and beat other RNN-based and CNN-based approaches which use human-define normalization methods \cite{CVPR15HRNN,Shahroudy_2016_CVPR} or use some advanced techniques \cite{liu2016spatio,AAAI17Atte,liu2017global,li2017RNNTree,liu2017enhanced}, such as attention \cite{AAAI17Atte,liu2017global}. Our proposed method, \emph{VA-fusion(aug.)}, outperforms the best state-of-the-art results by 9.4\% and 7.8\% on CS and CV settings, respectively.
\begin{table}[t] 
	\begin{center}
		\caption{Accuracy (\%) on the NTU dataset.} 
		\label{table:NTU}
		\begin{tabular}{c|c|c}
			\hline
			Method & CS  & CV \\
			\hline
			Skeleton Quads \cite{evangelidis2014skeletal}  & 38.6 & 41.4\\ 
			\hline			
			Lie Group \cite{vemulapalli2014human} & 50.1  & 52.8\\ 
			\hline
			Dynamic Skeletons  \cite{hu2015jointly} & 60.2  & 65.2\\
			\hline
			HBRNN-L  \cite{CVPR15HRNN} & 59.1 &  64.0\\
			\hline
			Part-aware LSTM   \cite{Shahroudy_2016_CVPR} & 62.9 & 70.3  \\
			\hline
			ST-LSTM + Trust Gate \cite{liu2016spatio} & 69.2 & 77.7  \\
			\hline
			STA-LSTM \cite{AAAI17Atte} & 73.4 &  81.2 \\ 
			\hline
			GCA-LSTM \cite{liu2017global} & 74.4 & 82.8 \\
			\hline
			URNN-2L-T \cite{li2017RNNTree} & 74.6 & 83.2 \\
			\hline
			Clips+CNN+MTLN \cite{ke2017new} & 79.6 & 84.8 \\
			\hline
			ESV~(Synthesized + Pre-trained) \cite{liu2017enhanced} & 80.0  & 87.2 \\				
			\hline
			\hline
			VA-RNN(aug.) & {79.8} & {88.9}\\ \hline
			VA-CNN(aug.) & {88.7} & {94.3} \\   \hline
			VA-fusion(aug.) & \textbf{89.4} & \textbf{95.0} \\
			\hline
		\end{tabular}
	\end{center}
	\vspace{-2.5mm}	
\end{table}

\subsubsection{SYSU Dataset}
We follow the standard protocols proposed by \cite{hu2015jointly} to evaluate the performance. For setting 1, half of the subjects are used for training and the others for testing. For setting 2, half of the videos of each subject are used for training and the others for testing. For each setting, the averaged results from 30-fold cross validation are shown in Table \ref{table:SYSU}. With view adaptation models reducing the view variations, our approach achieves the best performance, which is 11.2\% and 9.3\% higher than that of \cite{hu2015jointly} for setting-1 and setting-2, respectively, and 10.2\% higher than that of~\cite{Liu2016} for setting-1.

\begin{table}[t] 
	\begin{center}
		\caption{Accuracy (\%) on the SYSU dataset.}
		\label{table:SYSU}
		\begin{tabular}{c|c|c}
			\hline
			Method & setting-1  & setting-2 \\
			\hline
			LAFF \cite{hu2016real}  & 54.2 &--\\
			\hline
			Dynamic Skeletons \cite{hu2015jointly} & 75.5 & 76.9 \\	
			\hline
			ST-LSTM + Trust Gate \cite{Liu2016} & 76.5 & -- \\				
			\hline
			\hline
			VA-RNN(aug.)  & {80.5} & {79.8} \\
			\hline
			VA-CNN(aug.)  & {85.1} & {84.8} \\
			\hline
			VA-fusion(aug.)  & \textbf{86.7} & \textbf{86.2} \\
			\hline
		\end{tabular}
	\end{center}
\end{table}

\begin{table*}[htbp]
	\centering
	\caption{Accuracy (\%) on the UWA3D dataset.}
	\label{table:UWA3D}
	\begin{tabular}{c|c|c|c|c|c|c|c|c|c|c|c|c|c}
		\hline
		Training views &\multicolumn{2}{|c|}{V1 \& V2}     &\multicolumn{2}{|c|}{V1 \& V3}      &\multicolumn{2}{|c|}{V1 \& V4}      &\multicolumn{2}{|c|}{V2 \& V3}     &\multicolumn{2}{|c|}{V2 \& V4}     &\multicolumn{2}{|c|}{V3 \& V4}       & \multirow{2}{*}{Average} \\ \cline{1-13}
		Testing views & V3       & V4   & V2       & V4   & V2       & V3   & V1       & V4   & V1       & V3   & V1       & V2   &      \\ \hline
		HOJ3D \cite{xia2012view}         & 15.3     & 28.2 & 17.3     & 27.0 & 14.6     & 13.4 & 15.0     & 12.9 & 22.1     & 13.5 & 20.3     & 12.7 & 17.7 \\ \hline
		AE \cite{wang2014learning}           & 45.0     & 40.4 & 35.1     & 36.9 & 34.7     & 36.0 & 49.5     & 29.3 & 57.1     & 35.4 & 49.0     & 29.3 & 39.8 \\ \hline
		LARP \cite{vemulapalli2014human}          & 49.4     & 42.8 & 34.6     & 39.7 & 38.1     & 44.8 & 53.3     & 33.5 & 53.6     & 41.2 & 56.7     & 32.6 & 43.4 \\ \hline
		ESV~(Synthesized + Pre-trained) \cite{liu2017enhanced}       & 72.3     & 76.3 & 64.7     & 75.5 & 63.5     & 74.0 & 83.1     & 75.1 & 82.4     & 71.1 & 83.5     & 63.5 & 73.8 \\ \hline \hline
		VA-RNN(aug.)       & {70.9}     & 73.2 & {68.1}     & {72.0} & {68.1}     & {71.3} & {81.3}     & {76.8} & {79.4}     & {71.7} & {79.4}     & {71.3} & {73.6} \\ \hline
		VA-CNN(aug.)       & {77.7}     & 83.1 & {77.2}     & {83.1} & {73.2}     & {73.3} & {86.8}     & {79.9} & {84.8}     & {73.7} & {83.3}     & {75.2} & {79.3} \\ \hline
		VA-fusion(aug.)       & \textbf{80.9}     & \textbf{84.3} & \textbf{78.7}     & \textbf{86.2} & \textbf{75.2}     & \textbf{73.3} & \textbf{87.6}     & \textbf{84.3} & \textbf{86.0}     & \textbf{74.9} & \textbf{86.4}     & \textbf{79.5} & \textbf{81.4} \\ \hline
	\end{tabular}
\end{table*}

\subsubsection{UWA3D Dataset}

We follow the standard protocol proposed by \cite{rahmani2016histogram} to evaluate the performance. There are 4 views. The dataset is partitioned in different manners to have 12 kinds of partitions. Each partition has 3 viewpoints, where two are used for training and the other one is used for testing. Table \ref{table:UWA3D} shows the results on each partition. The four views are rather different which make it hard to recognize actions under unseen views. With the view adaptation model, \emph{VA-RNN(aug.)} and \emph{VA-CNN(aug.)} outperform the baseline schemes significantly. Our single model \emph{VA-CNN(aug.)} outperforms \emph{ESV} \cite{liu2017enhanced} largely by 5.5\% even though \emph{ESV} fuses 10 different models.

\subsubsection{N-UCLA Dataset}
\label{subsec:UCLA}

There are three views in this dataset. Usually, two of the views are used for training and the other one is used for testing \cite{wang2014cross,du2016representation}. In \cite{wang2014cross}, they only use samples of the first two views as training and the other one as testing. In \cite{du2016representation}, they choose every two views as training with 3 cases in total. Table \ref{table:UCLA} shows the comparisons of the performance. \emph{V1} denotes the partition that samples from views 2 and 3 are taken as the training samples and samples from view 1 as the testing samples. Similarly, \emph{V2} denotes samples of view 2 are taken as testing samples. With view adaptation modules, our scheme \emph{VA-fusion(aug.)} achieves the best performance of 95.3\% for the \emph{V3} setting. 

\begin{table}[t] 
	\centering
	\caption{Accuracy (\%) on the N-UCLA dataset.}
	\vspace{-2mm}
	\label{table:UCLA}
	\resizebox{0.45\textwidth}{!}{
	\begin{tabular}{c|c|c|c|c}
		\hline
		Setting~(test view) & V3 & V2 & V1 & Average \\ 
		\hline
		HOJ3D \cite{xia2012view}          & 54.5     & -        & -        &-      \\ \hline
		AE  \cite{wang2014learning}            & 76.0     & -        & -        &-      \\ \hline
		LARP \cite{vemulapalli2014human}           & 74.2     & -        & -        &-      \\ \hline
		HBRNN-L \cite{du2016representation}        & 78.5     & 83.5     & 79.3     & 80.5 \\ \hline
		ESV(Synthesized+Pre-trained)\cite{liu2017enhanced}          & 92.6     & -        & -        & -    \\ \hline
		\hline
		VA-RNN(aug.)         & {90.7}     & {87.5}     & {74.0}     & {84.1} \\ \hline
		VA-CNN(aug.)         & {93.8}     & {86.3}     & {79.7}     & {86.6} \\ \hline
		VA-fusion(aug.)         & \textbf{95.3}     & \textbf{88.7}     & \textbf{80.2}     & \textbf{88.1} \\ \hline
	\end{tabular}}
\end{table}

\subsubsection{SBU Dataset}
We follow the standard protocol proposed by \cite{yun2012two} which uses cross validation with 5 folders. Table \ref{table:SBU} shows the performance comparisons.
Our method outperforms other approaches \cite{CVPR15HRNN,zhu2015co,AAAI17Atte,liu2016spatio} significantly, achieving 98.3\% in accuracy. Even though there is no large view change in this dataset, our model tends to catch the slight view differences and transforms the views to suitable ones for more efficient action recognition. For this small dataset (only 282 sequences),  \emph{VA-RNN} performs better than \emph{VA-CNN}. That is because number of parameters of \emph{VA-CNN} is much larger than that of \emph{VA-RNN}, and it is easy for ConvNet to be overfitting for small training dataset.
\begin{table}[t] 
	\begin{center}
		\caption{Accuracy (\%) on the SBU dataset.} 
		\label{table:SBU}
		\begin{tabular}{c|c}
			\hline
			Methods & Acc. (\%) \\
			\hline
			Raw skeleton \cite{yun2012two} & 49.7 \\
			\hline
			Joint feature \cite{yun2012two} & 80.3 \\
			\hline
			Raw skeleton \cite{ji2014interactive} & 79.4 \\
			\hline
			Joint feature \cite{ji2014interactive} & 86.9 \\
			\hline
			HBRNN-L \cite{CVPR15HRNN} & 80.4 \\
			\hline
			Co-occurrence RNN \cite{zhu2015co} & 90.4 \\
			\hline
			STA-LSTM \cite{AAAI17Atte} & 91.5 \\
			\hline
			ST-LSTM + Trust Gate \cite{liu2016spatio} & 93.3 \\	
			\hline
			GCA-LSTM \cite{liu2017global} & 94.1 \\	
			\hline
			Clips+CNN+MTLN \cite{ke2017new} & 93.6 \\					
			\hline
			\hline	
			VA-RNN(aug.) & {97.5} \\ \hline
			VA-CNN(aug.) & {95.7} \\ \hline
			VA-fusion(aug.) & \textbf{98.3} \\
			\hline
		\end{tabular}
	\end{center}
\end{table}

Optimized with the target of maximizing the recognition performance, the proposed view adaptation model is very effective in choosing the suitable viewpoints. The consistency of viewpoints for various actions/subjects overcomes the challenge caused by the diversity of viewpoints in video capturing, enabling the network to focus on the learning of action-specific features. Unlike many pre-processing strategies, valuable motion information is preserved.    

\subsection{Comparative Analysis of VA-RNN and VA-CNN}
\label{subsec:complexity}

\begin{table*}[htbp]
	\centering
	\caption{Effectiveness (in accuracy(\%)) of the view adaptation design on different backbone CNN networks. Note that \emph{S-trans+CNN}(aug.) denotes the baseline scheme where sequence level translation pre-processing and data augmentation is performed. }
	\begin{tabular}{c|c|c|c|c|c|c}
		\hline
		\multicolumn{1}{c|}{Networks} & Method & \multicolumn{1}{c|}{\#Param.(M)} & \multicolumn{1}{c|}{CS} & \multicolumn{1}{c|}{CV} & \multicolumn{1}{c|}{CS gain} & \multicolumn{1}{c}{CV gain} \\
		\hline
		\multirow{2}[4]{*}[1mm]{CNN-6layers} & S-trans+CNN(aug.) & 1.93  & 79.6  & 85.1  & \multirow{2}[4]{*}[1mm]{1.4} & \multirow{2}[4]{*}[1mm]{3.3} \\
		\cline{2-5}          & VA-CNN(aug.) & 2.06  & 81.0    & 88.4  &       &  \\
		\hline
		\multirow{2}[4]{*}[1mm]{ResNet10} & S-trans+CNN(aug.) & 4.94  & 81.6 & 86.6  & \multirow{2}[4]{*}[1mm]{1.4} & \multirow{2}[4]{*}[1mm]{3.3} \\
		\cline{2-5}          & VA-CNN(aug.) & 5.39  & 83.0    & 89.9  &       &  \\
		\hline
		\multirow{2}[4]{*}[1mm]{ResNet18} &S-trans+CNN(aug.) & 11.21  & 86.5  & 93.1  & \multirow{2}[4]{*}[1mm]{1.2} & \multirow{2}[4]{*}[1mm]{0.8} \\
		\cline{2-5}          & VA-CNN(aug.) & 11.66 & 87.7  & 93.9  &       &  \\
		\hline
		\multirow{2}[4]{*}[1mm]{ResNet50} & S-trans+CNN(aug.) & 23.63 & 87.9  & 93.5  &\multirow{2}[4]{*}[1mm]{0.8}& \multirow{2}[4]{*}[1mm]{0.8} \\
		\cline{2-5}          & VA-CNN(aug.) & 24.09 & 88.7 & 94.3  &       &  \\
		\hline
	\end{tabular}%
	\label{tab:deep}%
\end{table*}%

\begin{table*}[!]
	\centering
	\caption{Effectiveness (in accuracy(\%)) of the view adaptation design with small network CNN-6layers and large network ResNet50 as the backnone CNN Networks. \emph{Gain} denotes the gap between \emph{S-trans+CNN}(aug.) and \emph{VA-CNN}(aug.). }
		\begin{tabular}{c|c|c|c|c|c|c|c|c}
			\hline
			\multirow{2}[4]{*}[1mm]{Network} & \multirow{2}[4]{*}[1mm]{Method} & \multicolumn{2}{c|}{NTU} & \multicolumn{2}{c|}{SYSU} & \multirow{2}[4]{*}[1mm]{UWA3D} & \multirow{2}[4]{*}[1mm]{N-UCLA} & \multirow{2}[4]{*}[1mm]{SBU} \\
			\cline{3-6}          &       & CS    & CV    & setting-1 & setting-2 &       &       &  \\
			\hline
			\multirow{3}[6]{*}[2mm]{ResNet50} &S-trans+CNN(aug.) &87.9  & 93.5  & 84.2  & 83.4  & 77.0  & 85.7  & 93.0  \\
			\cline{2-9}          & VA-CNN(aug.) & 88.7  & 94.3 & 85.1  & 84.8  & 79.3  & 86.6  & 95.7  \\
			\cline{2-9}          & Gain  & 0.8   & 0.8   & 0.9   & 1.4   & 2.3   & 0.9   & 2.7  \\
			\hline
			\multirow{3}[6]{*}[2mm]{CNN-6layer} & S-trans+CNN(aug.) & 79.6  & 85.1  & 74.6  & 74.4  & 55.8  & 72.0  & 82.7  \\
			\cline{2-9}          & VA-CNN(aug.) & 81.0  & 88.4  & 76.2  & 76.2  & 67.3  & 79.3  & 86.2  \\
			\cline{2-9}          & Gain &\textbf{ 1.4  } & \textbf{3.3 }  &\textbf{ 1.6}   & \textbf{1.8 }  & \textbf{11.5 } & \textbf{7.3}   &\textbf{3.5 } \\
			\hline
	\end{tabular}%
	\label{tab:cnn_gain_small_model}%
\end{table*}%

Thanks to the introduction of the view adaptation modulesxx, both \emph{VA-RNN}  and \emph{VA-CNN}  achieve improvement in comparison with their baselines, as shown in Table \ref{table:all-basline-ours}.

\emph{VA-CNN}(aug.) is much more powerful than \emph{VA-RNN}(aug.) in general as shown in Table \ref{table:all-basline-ours}. The main reason is that we transform the entire skeleton sequence to an image and deep CNN network (such as ResNet50) is able to explore the spatial and temporal relationship of the joints locally and globally. But RNN has limited memory of the history information.
	
The gain of the view adaptation module over deep CNN networks seems smaller than that of RNN networks. We conduct experiments on several backbone CNNs with different model sizes to analyze the effectiveness of view adaptation model. We take \emph{CNN-6layers}, which consists of five convolutional layers and one fc layer for classification, \emph{ResNet10}, \emph{ResNet18}, and \emph{ResNet50} as our backbone networks and show the results in Table \ref{tab:deep}. We have two conclusions. (1)  Our model achieves large gains when the CNNs are small. For the \emph{CNN-6layers} network, our view adaptation model achieves 1.4\% and 3.3\% gains on the CS and CV settings of the NTU dataset, which are comparable with the gains of RNN-based network as shown in Table \ref{table:all-basline-ours}. More results on all datasets are shown in Table \ref{tab:cnn_gain_small_model}. We can find that our view adaptation models achieve significant gains for all datasets compared to CNN baselines when the networks are small. (2) Our view adaptation module achieves sizable gains when the backbone CNNs are large. As the model size or complexity increases, it becomes harder to get the same gain. From Table \ref{tab:deep}, we see that the larger the model size of baseline ({\it{i.e.}} the number of the parameters from the model), the smaller gain achieved by increasing the model size by the same amount. However, with negligible increase in model size of the view adaptation module, the performance improves much more compared to increasing the depth of network.

In addition, we also shows the comparisons of our final \emph{VA-RNN}(aug.) and \emph{VA-CNN}(aug.) models  in terms of the number of parameters of the models, testing speed when the bach size is set to 1 (number of sequences per second), and the accuracy (\%) in Table \ref{table:models}. Note that the performance of deep CNN is superior to RNN of three LSTM layers. When more layers is utilized in RNN, the performance increase little. Here we assume the sequence length is 300 frames. (1) \emph{VA-RNN}(aug.) has the advantages of small model size (number of parameter), which is only 2\% of the model size of \emph{VA-CNN}(aug.) . (2) For the recognition on the well trimmed sequences, \emph{VA-CNN}(aug.) has rather high recognition speed, which is 83.3 sequences per second, about 10 times faster than \emph{VA-RNN(aug.)}. For the online detection task, considering the LSTM structure is suitable for frame-wise processing while \emph{VA-CNN}(aug.) needs to use a sliding window strategy to process the untrimmed streaming data, \emph{VA-RNN}(aug.) may be more time efficient, depending on the size of sliding window. For example, if the window slides for each frame, the speed of \emph{VA-CNN}(aug.) is about 83.3 frames/second while the speed of \emph{VA-RNN}(aug.) is about 7.9$\times$300 = 2370 frames/second. (3) \emph{VA-CNN}(aug.) has higher recognition accuracy than that of \emph{VA-RNN}(aug.) due to its joint spatio-temporal exploration capability, the power of the CNN structure, and larger model size. But for small dataset, \emph{VA-RNN}(aug.) has superior performance thanks to its small number of parameters. Depending on the requirement of practical applications, users can choose among \emph{VA-RNN}(aug.), \emph{VA-CNN}(aug.), and \emph{VA-fusion}(aug.).       

\begin{table}[t] 
	\begin{center}
		\caption{Model comparisons of VA-RNN and VA-CNN.} 
		\label{table:models}
		\resizebox{0.48\textwidth}{!}{
			\begin{tabular}{c|c|c|c}
				\hline
				Model & $\#$Param.(M)  & Speed~(seq./sec.) & Acc.(NTU-CV)(\%)\\
				\hline
				VA-RNN(aug.) & 0.47 & 7.9 & 88.7\\
				\hline
				VA-CNN(aug.) & 24.09 & 83.3 & 94.3\\	
				\hline
			\end{tabular}}
		\end{center}
		\vspace{-6mm}	
	\end{table}
\section{Conclusion}

We present two end-to-end view adaptive neural networks, VA-RNN and VA-CNN, for human action recognition from skeleton data. Rather than following the human predefined criterion to re-position skeletons for action recognition, the proposed networks are capable of adapting the observation viewpoints to the suitable ones by itself, with the optimization target of maximizing the recognition performance. We have designed view adaptation models based on the recurrent neural network and the convolutional neural network respectively. Both models can automatically transform the skeletons to consistent viewpoints which eliminate the influence of the diversity of viewpoints and ease the training. Experimental results demonstrate that the proposed framework consistently improves the recognition performance on five challenging benchmark datasets and achieves state-of-the-art performances.

%

\ifCLASSOPTIONcompsoc
  \section*{Acknowledgments}
\else
   regular IEEE prefers the singular form
  \section*{Acknowledgment}
\fi
Jianru Xue is supported by  National Key Research and Development Program of China under Grant 2016YFB1001004, Natural Science Foundation of China under Grant 61773311, and Grant 61751308. Junliang Xing is supported by the Natural Science Foundation of China (Grant No. 61672519).
%
%
\ifCLASSOPTIONcaptionsoff
  \newpage
\fi



%
%
%

\bibliographystyle{abbrv}
\bibliography{egbib}

\begin{thebibliography}{10}

\bibitem{IntelRealSense}
{Intel} {RealSense}.
\newblock \url{https://software.intel.com/en-us/realsense}.

\bibitem{aggarwal2014human}
J.~K. Aggarwal and L.~Xia.
\newblock Human activity recognition from 3d data: A review.
\newblock {\em Pattern Recognit. Lett.}, 48:70--80, 2014.

\bibitem{bashir2006feature}
F.~I. Bashir, A.~A. Khokhar, and D.~Schonfeld.
\newblock View-invariant motion trajectory-based activity classification and
  recognition.
\newblock {\em Multimedia Syst.}, 12(1):45--54, 2006.

\bibitem{du2015skeleton}
Y.~Du, Y.~Fu, and L.~Wang.
\newblock Skeleton based action recognition with convolutional neural network.
\newblock In {\em ACPR}. IEEE, 2015.

\bibitem{du2016representation}
Y.~Du, Y.~Fu, and L.~Wang.
\newblock Representation learning of temporal dynamics for skeleton-based
  action recognition.
\newblock {\em TIP}, 25(7):3010--3022, 2016.

\bibitem{CVPR15HRNN}
Y.~Du, W.~Wang, and L.~Wang.
\newblock Hierarchical recurrent neural network for skeleton based action
  recognition.
\newblock In {\em CVPR}, 2015.

\bibitem{evangelidis2014skeletal}
G.~Evangelidis, G.~Singh, and R.~Horaud.
\newblock Skeletal quads: Human action recognition using joint quadruples.
\newblock In {\em ICPR}, 2014.

\bibitem{Farhadi2008Wrongview}
A.~Farhadi and M.~K. Tabrizi.
\newblock Learning to recognize activities from the wrong view point.
\newblock In {\em ECCV}, 2008.

\bibitem{feng2015usemoreview}
J.-g. Feng and X.~Jun.
\newblock View-invariant human action recognition via robust locally adaptive
  multi-view learning.
\newblock {\em Frontiers of Information Technology \& Electronic Engineering},
  16(11):917--920, 2015.

\bibitem{han2016space}
F.~Han, B.~Reily, W.~Hoff, and H.~Zhang.
\newblock Space-time representation of people based on 3d skeletal data: A
  review.
\newblock {\em CVIU}, 158:85--105, 2017.

\bibitem{he2016deep}
K.~He, X.~Zhang, S.~Ren, and J.~Sun.
\newblock Deep residual learning for image recognition.
\newblock In {\em CVPR}, 2016.

\bibitem{hu2015jointly}
J.-F. Hu, W.-S. Zheng, J.~Lai, and J.~Zhang.
\newblock Jointly learning heterogeneous features for {RGB-D} activity
  recognition.
\newblock In {\em CVPR}, 2015.

\bibitem{hu2016real}
J.-F. Hu, W.-S. Zheng, L.~Ma, G.~Wang, and J.~Lai.
\newblock Real-time {RGB-D} activity prediction by soft regression.
\newblock In {\em ECCV}, 2016.

\bibitem{iosifidis2012ANN}
A.~Iosifidis, A.~Tefas, and I.~Pitas.
\newblock View-invariant action recognition based on artificial neural
  networks.
\newblock {\em TNNLS}, 23(3):412--424, 2012.

\bibitem{ji2010advances}
X.~Ji and H.~Liu.
\newblock Advances in view-invariant human motion analysis: A review.
\newblock {\em TSMCC}, 40(1):13--24, 2010.

\bibitem{ji2014interactive}
Y.~Ji, G.~Ye, and H.~Cheng.
\newblock Interactive body part contrast mining for human interaction
  recognition.
\newblock In {\em ICMEW}, 2014.

\bibitem{jiang2015informative}
M.~Jiang, J.~Kong, G.~Bebis, and H.~Huo.
\newblock Informative joints based human action recognition using skeleton
  contexts.
\newblock {\em Signal Processing: Image Communication}, 33:29--40, 2015.

\bibitem{PP73Perception}
G.~Johansson.
\newblock Visual perception of biological motion and a model for it is
  analysis.
\newblock {\em Perception and Psychophysics}, 14(2):201--211, 1973.

\bibitem{junejo2008selfsimilarities}
I.~N. Junejo, E.~Dexter, I.~Laptev, and P.~P{\'e}rez.
\newblock Cross-view action recognition from temporal self-similarities.
\newblock In {\em ECCV}, 2008.

\bibitem{ke2017new}
Q.~Ke, M.~Bennamoun, S.~An, F.~Sohel, and F.~Boussaid.
\newblock A new representation of skeleton sequences for 3d action recognition.
\newblock In {\em CVPR}, pages 601--604, 2017.

\bibitem{kingma2014adam}
D.~P. Kingma and J.~Ba.
\newblock Adam: A method for stochastic optimization.
\newblock {\em arXiv:1412.6980}, 2014.

\bibitem{krizhevsky2012imagenet}
A.~Krizhevsky, I.~Sutskever, and G.~E. Hinton.
\newblock Imagenet classification with deep convolutional neural networks.
\newblock In {\em Advances in neural information processing systems}, pages
  1097--1105, 2012.

\bibitem{li2017skeleton}
B.~Li, Y.~Dai, X.~Cheng, H.~Chen, Y.~Lin, and M.~He.
\newblock Skeleton based action recognition using translation-scale invariant
  image mapping and multi-scale deep cnn.
\newblock In {\em Multimedia \& Expo Workshops (ICMEW), 2017 IEEE International
  Conference on}, pages 601--604. IEEE, 2017.

\bibitem{li2012virtualview}
R.~Li and T.~Zickler.
\newblock Discriminative virtual views for cross-view action recognition.
\newblock In {\em CVPR}, 2012.

\bibitem{li2017RNNTree}
W.~Li, L.~Wen, M.-C. Chang, S.~N. Lim, and S.~Lyu.
\newblock Adaptive rnn tree for large-scale human action recognition.
\newblock In {\em ICCV}, 2017.

\bibitem{liu2011knowtransfer}
J.~Liu, M.~Shah, B.~Kuipers, and S.~Savarese.
\newblock Cross-view action recognition via view knowledge transfer.
\newblock In {\em ICCV}, 2011.

\bibitem{liu2016spatio}
J.~Liu, A.~Shahroudy, D.~Xu, and G.~Wang.
\newblock Spatio-temporal {LSTM} with trust gates for {3D} human action
  recognition.
\newblock {\em ECCV}, 2016.

\bibitem{Liu2016}
J.~Liu, A.~Shahroudy, D.~Xu, and G.~Wang.
\newblock Spatio-temporal lstm with trust gates for 3d human action
  recognition.
\newblock In {\em ECCV}, 2016.

\bibitem{liu2017global}
J.~Liu, G.~Wang, P.~Hu, L.-Y. Duan, and A.~C. Kot.
\newblock Global context-aware attention lstm networks for 3d action
  recognition.
\newblock In {\em CVPR}, 2017.

\bibitem{liu2017enhanced}
M.~Liu, H.~Liu, and C.~Chen.
\newblock Enhanced skeleton visualization for view invariant human action
  recognition.
\newblock {\em PR}, 68:346--362, 2017.

\bibitem{mahasseni2013latent}
B.~Mahasseni and S.~Todorovic.
\newblock Latent multitask learning for view-invariant action recognition.
\newblock In {\em ICCV}, 2013.

\bibitem{IVC10SurveyAction}
R.~Poppe.
\newblock A survey on vision-based human action recognition.
\newblock {\em IVC}, 28(6):976--990, 2010.

\bibitem{presti20163d}
L.~L. Presti and M.~La~Cascia.
\newblock 3d skeleton-based human action classification: a survey.
\newblock {\em PR}, 53:130--147, 2016.

\bibitem{rahmani2017learning}
H.~Rahmani and M.~Bennamoun.
\newblock Learning action recognition model from depth and skeleton videos.
\newblock In {\em ICCV}, 2017.

\bibitem{rahmani2016histogram}
H.~Rahmani, A.~Mahmood, D.~Huynh, and A.~Mian.
\newblock Histogram of oriented principal components for cross-view action
  recognition.
\newblock {\em TPAMI}, 38(12):2430--2443, 2016.

\bibitem{rahmani2015knotransfer}
H.~Rahmani and A.~Mian.
\newblock Learning a non-linear knowledge transfer model for cross-view action
  recognition.
\newblock In {\em CVPR}, 2015.

\bibitem{rao2001view}
C.~Rao and M.~Shah.
\newblock View-invariance in action recognition.
\newblock In {\em CVPR}, 2001.

\bibitem{Shahroudy_2016_CVPR}
A.~Shahroudy, J.~Liu, T.-T. Ng, and G.~Wang.
\newblock {NTU RGB+D}: A large scale dataset for {3D} human activity analysis.
\newblock In {\em CVPR}, 2016.

\bibitem{shen2008ratio}
Y.~Shen and H.~Foroosh.
\newblock View-invariant action recognition using fundamental ratios.
\newblock In {\em CVPR}, 2008.

\bibitem{shen2009pointtriplets}
Y.~Shen and H.~Foroosh.
\newblock View-invariant action recognition from point triplets.
\newblock {\em TPAMI}, 31(10):1898--1905, 2009.

\bibitem{CVPR11BestPaper}
J.~Shotton, A.~Fitzgibbon, M.~Cook, T.~Sharp, M.~Finocchio, R.~Moore,
  A.~Kipman, and A.~Blake.
\newblock Real-time human pose recognition in parts from single depth images.
\newblock In {\em CVPR}, 2011.

\bibitem{AAAI17Atte}
S.~Song, C.~Lan, J.~Xing, W.~Zeng, and J.~Liu.
\newblock An end-to-end spatio-temporal attention model for human action
  recognition from skeleton data.
\newblock In {\em AAAI}, pages 4263--4270, 2017.

\bibitem{srivastava2014dropout}
N.~Srivastava, G.~E. Hinton, A.~Krizhevsky, I.~Sutskever, and R.~Salakhutdinov.
\newblock Dropout: a simple way to prevent neural networks from overfitting.
\newblock {\em JMLR}, 15(1):1929--1958, 2014.

\bibitem{sutskever2014sequence}
I.~Sutskever, O.~Vinyals, and Q.~V. Le.
\newblock Sequence to sequence learning with neural networks.
\newblock In {\em NIPS}, pages 3104--3112.

\bibitem{vemulapalli2014human}
R.~Vemulapalli, F.~Arrate, and R.~Chellappa.
\newblock Human action recognition by representing 3{D} skeletons as points in
  a lie group.
\newblock In {\em CVPR}, pages 588--595.

\bibitem{wang2014learning}
J.~Wang, Z.~Liu, and Y.~Wu.
\newblock Learning actionlet ensemble for 3d human action recognition.
\newblock In {\em Human Action Recognition with Depth Cameras}, pages 11--40.
  2014.

\bibitem{wang2014cross}
J.~Wang, X.~Nie, Y.~Xia, Y.~Wu, and S.-C. Zhu.
\newblock Cross-view action modeling, learning and recognition.
\newblock In {\em CVPR}, 2014.

\bibitem{wang2016temporal}
L.~Wang, Y.~Xiong, Z.~Wang, Y.~Qiao, D.~Lin, X.~Tang, and L.~Van~Gool.
\newblock Temporal segment networks: Towards good practices for deep action
  recognition.
\newblock In {\em ECCV}, pages 20--36.

\bibitem{wang2016action}
P.~Wang, Z.~Li, Y.~Hou, and W.~Li.
\newblock Action recognition based on joint trajectory maps using convolutional
  neural networks.
\newblock In {\em ACM MM}, 2016.

\bibitem{weinland2010making}
D.~Weinland, M.~{\"O}zuysal, and P.~Fua.
\newblock Making action recognition robust to occlusions and viewpoint changes.
\newblock In {\em ECCV}, 2010.

\bibitem{CVIU11SurveyAction}
D.~Weinland, R.~Ronfard, and E.~Boyerc.
\newblock A survey of vision-based methods for action representation,
  segmentation and recognition.
\newblock {\em CVIU}, 115(2):224--241, 2011.

\bibitem{wu2012latentSVM}
X.~Wu and Y.~Jia.
\newblock View-invariant action recognition using latent kernelized structural
  svm.
\newblock In {\em ECCV}, 2012.

\bibitem{wu2013cross}
X.~Wu, H.~Wang, C.~Liu, and Y.~Jia.
\newblock Cross-view action recognition over heterogeneous feature spaces.
\newblock In {\em ICCV}, 2013.

\bibitem{xia2012view}
L.~Xia, C.-C. Chen, and J.~Aggarwal.
\newblock View invariant human action recognition using histograms of 3d
  joints.
\newblock In {\em CVPRW}, 2012.

\bibitem{CVPR12HO3DJ}
L.~Xia, C.-C. Chen, and J.~K. Aggarwal.
\newblock View invariant human action recognition using histograms of 3{D}
  joints.
\newblock In {\em CVPRW}.

\bibitem{yun2012two}
K.~Yun, J.~Honorio, D.~Chattopadhyay, T.~L. Berg, and D.~Samaras.
\newblock Two-person interaction detection using body-pose features and
  multiple instance learning.
\newblock In {\em CVPRW}, 2012.

\bibitem{zhang2016rgb}
J.~Zhang, W.~Li, P.~O. Ogunbona, P.~Wang, and C.~Tang.
\newblock Rgb-d-based action recognition datasets: A survey.
\newblock {\em PR}, 60:86--105, 2016.

\bibitem{zhang2017VA}
P.~Zhang, C.~Lan, J.~Xing, W.~Zeng, J.~Xue, and N.~Zheng.
\newblock View adaptive recurrent neural networks for high performance human
  action recognition from skeleton data.
\newblock In {\em ICCV}, pages 2117--2126.

\bibitem{zhang2012microsoft}
Z.~Zhang.
\newblock Microsoft kinect sensor and its effect.
\newblock {\em IEEE MultiMedia}, 19(2):4--10, 2012.

\bibitem{zhang2013virtualpath}
Z.~Zhang, C.~Wang, B.~Xiao, W.~Zhou, S.~Liu, and C.~Shi.
\newblock Cross-view action recognition via a continuous virtual path.
\newblock In {\em CVPR}, 2013.

\bibitem{zheng2013sparse}
J.~Zheng and Z.~Jiang.
\newblock Learning view-invariant sparse representations for cross-view action
  recognition.
\newblock In {\em ICCV}, 2013.

\bibitem{zhu2015co}
W.~Zhu, C.~Lan, J.~Xing, W.~Zeng, Y.~Li, L.~Shen, and X.~Xie.
\newblock Co-occurrence feature learning for skeleton based action recognition
  using regularized deep {LSTM} networks.
\newblock In {\em AAAI}, 2016.

\end{thebibliography}

\end{document}